\documentclass{article} % For LaTeX2e
\usepackage{iclr2027_conference,times}
\usepackage{amsmath,amsfonts,bm}

\def\eqref#1{equation~\ref{#1}}
\def\1{\bm{1}}

\DeclareMathAlphabet{\mathsfit}{\encodingdefault}{\sfdefault}{m}{sl}
\SetMathAlphabet{\mathsfit}{bold}{\encodingdefault}{\sfdefault}{bx}{n}

\usepackage{url}
\usepackage{multirow}
\usepackage{wrapfig}
\definecolor{mydarkred}{rgb}{0.6,0,0}
\definecolor{mydarkgreen}{rgb}{0,0.6,0}
\usepackage[colorlinks,
linkcolor=mydarkred,
citecolor=blue]{hyperref}
\usepackage{soul} % 引入 soul 包用于文字高亮
\usepackage{hyperref}
\usepackage{pifont}
\usepackage{url}
\usepackage[pdftex]{graphicx}
\usepackage{xspace}
\usepackage{booktabs}
\usepackage{amsmath}
\usepackage{amssymb}
\usepackage[table]{xcolor}  % 支持表格中的颜色命令，如 \rowcolor
\usepackage{subcaption} % 处理 subfigure 环境
\definecolor{fix}{RGB}{0, 102, 204}
\usepackage{cancel}
\usepackage{mathtools}
\usepackage[table]{xcolor}
\usepackage{booktabs}
\usepackage{makecell}
\usepackage{multirow}
\usepackage{algorithm}
\usepackage{algpseudocode}
\usepackage{tcolorbox}
\usepackage{amsthm}
\usepackage{wrapfig}
\usepackage{tcolorbox}
\tcbuselibrary{most}  % 启用 enhanced 功能
\usepackage{boxedminipage}
\usepackage{relsize}

\usepackage{minitoc}

\title{Understanding Evaluation Inconsistency in Diffusion Large Language Models}

\author{
Hengxiang Zhang\textsuperscript{1},\enspace
Jiaxi Ren \textsuperscript{1},\enspace
Renchunzi Xie \textsuperscript{2},\enspace
Hongxin Wei \textsuperscript{1}\thanks{Corresponding author} \\
\textsuperscript{1}Department of Statistics and Data Science, Southern University of Science and Technology \\
\textsuperscript{2} Intelligent Computing Research Center, Great Bay Institute for Advanced Study, Great Bay University \\
}

\newtcolorbox[auto counter]{examplebox}[2][]{
    colback=blue!5,
    colframe=gray!100,
    boxrule=0.5pt,
    arc=2pt,
    left=5pt,
    right=5pt,
    top=5pt,
    bottom=5pt,
    title={Example \thetcbcounter: #2},
    #1
}

\iclrfinalcopy % Uncomment for camera-ready version, but NOT for submission.

\newcommand{\up}[1]{\textcolor{green!50!black}{\scriptsize #1}}
\newcommand{\down}[1]{\textcolor{red}{\scriptsize #1}}
\newcommand{\score}[2]{\ensuremath{{#1}_{\raisebox{-0.3ex}{#2}}}}

\newtcolorbox[auto counter, number within=section]{promptbox}[2][]{
    breakable,
    title={Example~\thetcbcounter: #2},
    label={#1}
}
\title{\MakeLowercase{Understanding evaluation inconsistency in diffusion large language models}}

\begin{document}

\doparttoc % Tell to minitoc to generate a toc for the parts
\faketableofcontents % Run a fake tableofcontents command for the partocs
\maketitle

\begin{abstract}
Despite the capability of parallel decoding, diffusion large language models (dLLMs) require many denoising steps to maintain generation quality, motivating recent research on efficient decoding strategies.
However, existing studies have reported inconsistent benchmark results even in seemingly identical evaluation settings, undermining the reliability of conclusions about dLLM decoding.
To understand this issue, we perform a comprehensive empirical study of existing dLLM decoding methods across diverse evaluation protocols. 
Our analysis reveals that the performance gains of decoding methods are highly prompt-dependent, resulting in inconsistent rankings of methods across prompt templates.
We further attribute this evaluation inconsistency to the high sensitivity of parallel decoding methods to prompt design, where even minor prompt variations can significantly alter generation behavior.
Notably, we find that a well-chosen prompt template can yield markedly greater accuracy gains than simply increasing the decoding budget.
Beyond prompt templates, our experiments indicate that other commonly overlooked evaluation settings can also affect the assessment of decoding methods. 
Based on these findings, we provide practical guidelines for reliably evaluating decoding methods for diffusion large language models.
\end{abstract}

% \begin{abstract}
% Despite the capability of parallel decoding, diffusion large language models (dLLMs) require many denoising steps to maintain generation quality, motivating recent research on efficient decoding strategies.
% However, existing studies have reported inconsistent benchmark results even in seemingly identical evaluation settings, undermining the reliability of conclusions about dLLM decoding.
% To understand this inconsistency, we conduct a rigorous evaluation of current decoding methods for dLLMs across diverse evaluation settings.
% Surprisingly, our analysis reveals that the ranking of decoding methods is highly sensitive to the choice of prompt templates.
% Through comprehensive experiments across diverse prompt templates, we find that the performance gains of parallel decoding methods are highly prompt-dependent, and that aggressively accelerating generation can incur accuracy degradation on downstream tasks.
% We further attribute this inconsistency in evaluation to the high sensitivity of parallel decoding methods to minor variations in prompt templates.
% Notably, a well-chosen prompt template can yield markedly greater accuracy gains than simply increasing the decoding budget.
% Beyond prompt templates, our experiments indicate that other commonly overlooked evaluation settings can also affect the assessment of decoding methods.
% Based on these findings, we provide practical guidelines for reliably evaluating decoding methods for diffusion large language models.
% \end{abstract}

\section{Introduction}
Diffusion large language models (dLLMs) have emerged as a promising alternative to autoregressive large language models~\citep{nie2026large, ye2025dream, song2025seed}.
By adopting bidirectional attention, dLLMs enable any-order parallel decoding, overcoming the sequential generation bottleneck of autoregressive (AR) language models.
Despite the capability of parallel decoding, dLLMs still require a large number of denoising steps to maintain generation quality, considerably limiting their inference efficiency.
Recently, many studies have sought to accelerate dLLMs by leveraging key-value caches to reduce per-step computational cost, or by reducing the number of denoising steps~\citep{wu2026fastdllm, lu2026semanticaware, ma2025dkvcache}.
These methods have been reported to achieve substantial inference speedups without degrading generation quality, and in some cases even improve downstream performance.

However, benchmark scores reported in existing studies exhibit marked inconsistency even in seemingly identical evaluation settings, undermining the reliability of the evaluation of decoding methods~\citep{nie2026large, chen2026dparallel,liu2026lookahead}. 
For example, the accuracy reported on GSM8K for LLaDA-8B-Instruct exhibits a discrepancy of up to 9 percentage points across different studies, despite using the same decoding method and generation configuration~\citep{hu2025flashdlm, wu2026dynamicdllm}.
To investigate this inconsistency in evaluation, we performed a rigorous evaluation of current decoding methods across various benchmarks and models.
Our evaluation reveals that the performance of decoding methods is highly prompt-dependent.
This highlights a key pitfall in dLLM evaluation: the performance ranking of decoding methods can vary significantly across different prompt templates.
Through comprehensive evaluation across common prompt templates, we find that aggressive acceleration often incurs task accuracy degradation.
% Through comprehensive experiments across commonly used prompt templates, we find that aggressive acceleration generation often incurs task accuracy degradation.
% % Through comprehensive experiments across commonly used prompt templates, we find that the performance gains of parallel decoding methods are highly prompt-dependent and that aggressive acceleration often incurs task accuracy degradation.

\begin{figure}
% \vspace{-8mm}
    \centering
     \includegraphics[width=1\linewidth,height=0.19\textheight]{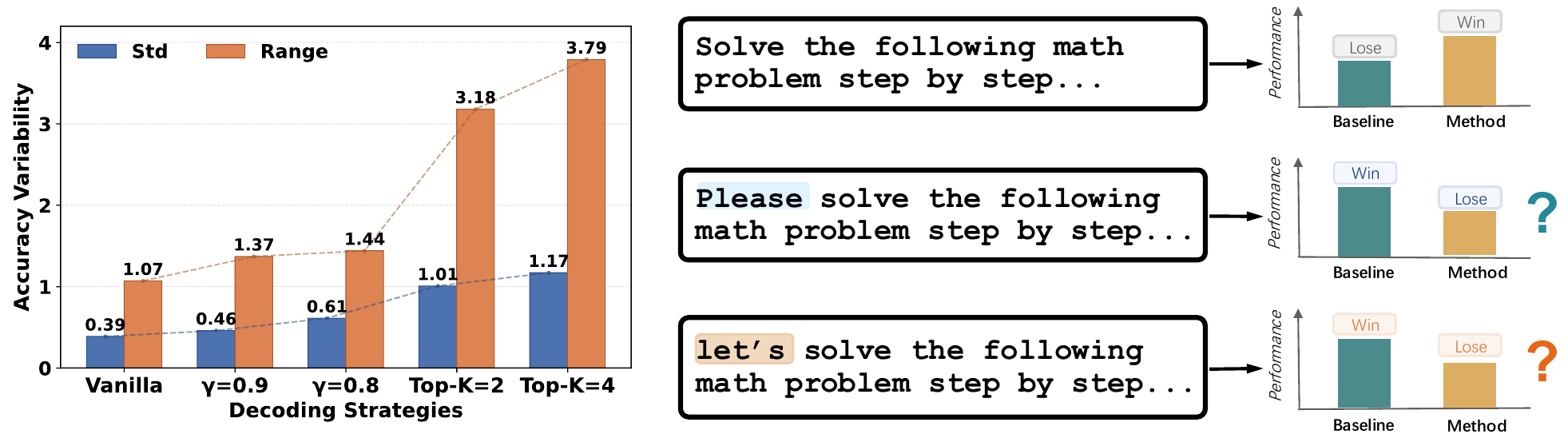}
    \caption{\textbf{Left}: Accuracy variability of single-token decoding (\textit{Vanilla}) and parallel decoding methods (\textit{Threshold-based decoding and Top-K}) across near-identical prompt templates. 
    We evaluate \textit{Threshold-based decoding} with $\gamma \in \{0.8, 0.9\}$ and \textit{Top-K decoding} with $k \in \{2,4\}$.
    Std and Range denote the standard deviation and range of accuracy across prompt templates. 
    \textbf{Right}: Minor variations in prompt templates can produce noticeable differences in performance, which may lead to misleading conclusions in method comparison.}
    \label{fig:abs}
\vspace{-5mm}
\end{figure}

In this work, we attribute this inconsistency in evaluation to the high sensitivity of decoding methods to minor prompt variations.
As illustrated in Figure~\ref{fig:abs}, minor variations in the prompt template can lead to non-trivial differences in downstream task performance.
This effect is markedly more pronounced for parallel decoding methods.
Specifically, parallel decoding is more prone to amplifying subtle prompt variations into inconsistent outputs, leading to performance fluctuations across prompts.
Moreover, our analysis shows that prompt templates can play a dominant role in determining evaluation results.
A well-chosen prompt template can yield markedly greater accuracy gains than simply increasing the decoding budget.
% An effective template can yield strong performance scores even with a smaller decoding budget, substantially exceeding the marginal gains from increasing the decoding budget.
% as simultaneously decoding multiple tokens violates the conditional independence assumption~\citep{kang2026parallelbench}, leading to dependency conflicts that further amplify performance variation.

Beyond prompt templates, our experiments indicate that commonly overlooked evaluation settings can also confound assessment of dLLM decoding methods. 
Specifically, increasing the generation length budget does not consistently yield performance gains and can even degrade performance under certain prompt templates.
In addition, few-shot evaluation settings generally improve accuracy.
However, decoding methods exhibit different preferences for the number of few-shot examples, making performance comparisons sensitive to the chosen few-shot setting.
Moreover, we find that evaluating decoding methods across different hardware platforms can yield inconsistent results, undermining fair comparisons.
Based on these findings, we highlight several practical considerations for reliable evaluation of decoding methods in dLLMs, such as multi-prompt evaluations and detailed reporting of evaluation settings. 

We summarize our main contributions as follows:
\begin{itemize}
    % \item We reveal a key evaluation pitfall in evaluating dLLM decoding methods: the performance rankings of decoding methods can vary significantly across prompt templates, undermining the reliability of evaluations.
    \item We reveal that the performance gains of decoding methods are highly prompt-dependent, resulting in inconsistent rankings of methods across prompt templates.
    \item We attribute this evaluation inconsistency to the high sensitivity of parallel decoding to prompt design, where even minor prompt variations can alter generation behavior.
    \item We provide a systematic analysis of dLLM decoding methods across various benchmarks and evaluation settings. Our analysis shows that, beyond prompt templates, commonly overlooked evaluation settings can affect evaluation results.
\end{itemize}

    % Parallel decoding methods are more susceptible to minor variations in prompt templates.

\section{Preliminary}
\subsection{Diffusion Large Language Models}
Diffusion large language models formulate text generation as a forward process and a reverse process. 
The forward process gradually adds noise to clean text by masking tokens, while the reverse process recovers masked sequences by iteratively predicting masked tokens.

\paragraph{Forward Process}
Let $x_0$ denote a clean sequence consisting of $N$ tokens.
We denote by $x_{t}$ the partially masked sequence, where each token is independently masked with probability $t \in [0,1]$. 
The forward process progressively masks tokens independently in $x_{0}$, resulting in a corrupted $x_t$. 
At each time $t$, the forward process is defined by the transition probability:
\begin{equation}
q_{t\mid 0}(x_t \mid x_0; t)=\prod_{i=1}^{N} q_{t\mid 0}(x_t^i \mid x_0^i),
\quad \text{where} \quad 
q_{t\mid 0}(x_t^i \mid x_0^i; t)=
\begin{cases}
t, & \text{if } x_t^i=[\texttt{MASK}],\\
1-t, & \text{if } x_t^i=x_0^i,\\
0, & \text{otherwise.}
\end{cases}
\end{equation}
where $q_{t\mid 0}(x_t^i \mid x_0^i; t)$ denotes the transition probability for the $i$-th token, where $x^i_t$ is independently replaced by [\texttt{MASK}] with probability $t$. 
The sequence-level transition probability 
$q_{t|0}(x_t \mid x_0; t)$ can be factorized as the product of token-level transition probabilities at each position.

\paragraph{Reverse Process}
The reverse process seeks to recover masked sequences by predicting masked tokens.
Specifically, the mask predictor is parameterized by a neural network $p_\theta$ that takes $x_{t}$ as input and predicts all masked tokens simultaneously. Formally, we define the reverse process as follows:

\begin{equation}
p_\theta\!\left(x_0 \mid x_{t};t \right)
=
\prod_{i\in \mathcal{M}_t}
p_\theta\!\left(x_0^{i} \mid x_t; t\right),
\end{equation}

where $\mathcal{M}_t:=\{i\mid x_t^i =[\texttt{MASK}]\}$ represents masked token set. The core of dLLMs is to learn a mask predictor that recovers masked tokens from corrupted text.
Consequently, given the noisy sequence, the model is trained to denoise $x_{t}$ by maximizing a variational lower bound on the log-likelihood:
\begin{equation}
\log p_\theta(x_0)
\ge
\mathbb{E}_{t, x_t}
\left[
\frac{1}{t}
\sum_{i=1}^{N}
\mathbf{1}[x_t^i=\texttt{MASK}]
\log p_{\mathrm{\theta}}(x_{0}^{i}\mid x_{t})
\right],
\end{equation}
where $t$ is sampled from the uniform distribution $t \sim U(0,1)$, and the $x_t$ is 
the corrupted text sampled from the distribution $q_{t\mid 0}(x_t \mid x_0; t)$.

\paragraph{Generation} At generation time, given a prompt $q$ with a fully masked response, the generation process progressively denoises the masked response into plain text over $T$ steps.
To maintain generation quality, the mask predictor decodes only a subset of the masked tokens at each step, rather than decoding all masked tokens simultaneously.
During generation, the model progressively decodes the masked response through iterative refinement over multiple steps.
As a result, the masked response is gradually unmasked through iterative refinement, resulting in a coherent and fluent output.
% \subsection{Parallel decoding strategies}

\section{Evaluation Inconsistency in Diffusion Large Language Models}
\subsection{Problem Setup}
Accelerating decoding in dLLMs typically degrades generation quality.
As a result, recent work has sought to develop decoding algorithms that improve inference efficiency without sacrificing generation quality.
To assess the performance of decoding methods, researchers commonly evaluate them on benchmark datasets, where higher accuracy indicates better generation quality.

Formally, we denote by $\mathcal{A} = \{A_1, \dots, A_n\}$ a set of decoding algorithms for dLLMs and $\mathcal{D}$ a benchmark dataset.
Given a decoding algorithm $A_i$, we query the model for each benchmark question $q$ with a prompt template $p$, and evaluate the response generated to obtain the performance score $s(A_i, p; \mathcal{D})$.
The performance scores of all decoding algorithms induce a ranking over $\mathcal{A}$:
\begin{equation}
s(A_{(1)},p;\mathcal{D})
\ge
s(A_{(2)},p;\mathcal{D})
\ge
\cdots
\ge
s(A_{(n)},p;\mathcal{D}),
\end{equation}
where $A_{(r)}$ denotes the decoding algorithm with the $r$-th highest score. 
A higher score often indicates that the decoding algorithm produces better performance on the benchmark dataset. 
This evaluation typically assumes that the relative ranking of decoding algorithms remains unchanged across different prompt templates.

\textbf{Definition 1} (\textit{Evaluation Inconsistency}).
Let $\mathcal{P}$ denote the set of prompt templates.
For each prompt template $p \in \mathcal{P}$, let $s(A_i, p)$ denote the performance score of method $A_i$ under prompt $p$.
The evaluation procedure exhibits \textit{evaluation inconsistency} if there exist $p, p' \in \mathcal{P}$ and $A_i, A_j \in \mathcal{A}$ such that
\[
s(A_i, p) > s(A_j, p) \quad \text{and} \quad s(A_i, p') < s(A_j, p').
\]
While performance scores may vary across prompt templates, a reliable evaluation is expected to produce a stable ranking of methods on the same benchmark.
Evaluation inconsistency implies that single-prompt evaluations yield high-variance estimates of method performance, leading to potentially biased performance assessments.

% Evaluation inconsistency implies that assessing decoding methods under a single prompt template may yield unreliable comparisons, leading to evaluation illusions.

\subsection{Evaluation Inconsistency Undermines Performance Metric Reliability}

\paragraph{Experimental setup}
We evaluate recent decoding methods for dLLMs, comparing Low-confidence~\citep{nie2026large} as a vanilla baseline against Fast-dLLM~\citep{wu2026fastdllm}, AdaBlock-dLLM~\citep{lu2026semanticaware}, AdaBlock-Fast~\citep{lu2026semanticaware}, dKV-Cache-Decode~\citep{ma2025dkvcache}, Elastic-Cache~\citep{nguyen-tri2026attention}, and WINO~\citep{hong2026widein}, on commonly used math and code benchmarks using LLaDA~\citep{zhullada, nie2026large} and Dream~\citep{ye2025dream} models.
In our experiments, we set the generation length to 128 and adopt the semi-autoregressive decoding strategy. 
\textit{To eliminate the effects of hardware and system configuration on evaluation results, our experiments are conducted on a single machine with a single GPU and a batch size of 1.}
For each decoding method, we use the best-performing hyperparameter settings reported in the corresponding paper.
Further details on the hyperparameter configurations are provided in Appendix~\ref{baselines}.
We evaluate decoding methods across prompts commonly used in existing evaluation practice, as presented in Appendix~\ref{label:benchmark_prompts}. 
In our experiments, \textit{Vanilla} denotes the low-confidence decoding baseline, which decodes a single masked token with the highest confidence at each step. 
Parallel decoding methods (e.g., Fast-dLLM) decode multiple masked tokens at each step according to a designed parallel decoding strategy.

\begin{table}[t]
\centering
\caption{The accuracy of vanilla and current decoding methods on GSM8K and MATH500 under LLaDA-8B-Instruct, across various prompt templates. \textit{Vanilla} baseline denotes the single-token decoding method. Each numerical subscript indicates the \textcolor{mydarkgreen}{improvement} or \textcolor{mydarkred}{degradation} compared to the vanilla decoding method.}
\label{tab:math_llada_results}
\renewcommand\arraystretch{1.5}
\resizebox{\textwidth}{!}{
\setlength{\tabcolsep}{0.8mm}{
\begin{tabular}{lllllllll} % |l for Avg.
\toprule
\multirow{2}{*}{Method} & \multicolumn{6}{c}{GSM8K} & \multicolumn{2}{c}{MATH500} % & \multicolumn{1}{c}{\multirow{2}{*}{Avg.}}
\\
\cmidrule(lr){2-7} \cmidrule(lr){8-9}
& Template 1 & Template 2 & Template 3 & Template 4 & Template 5 & Template 6 & Template 7 & Template 8 % &
\\
\midrule

\rowcolor{gray!10}
Vanilla
& \textbf{77.33} & \underline{74.60} & 71.72 & \textbf{76.35} & 74.98 & \underline{76.88} & \textbf{37.60} & 34.00 % & 65.43
\\
\midrule
\textit{Parallel decoding methods} \\
Fast-dLLM
& \score{76.35}{\down{-0.98}}
& \score{73.16}{\down{-1.44}}
& \score{70.51}{\down{-1.21}}
& \score{75.74}{\down{-0.61}}
& \score{\underline{75.82}}{\up{+0.84}}
& \score{74.60}{\down{-2.28}}
& \score{33.80}{\down{-3.80}}
& \score{\underline{35.20}}{\up{+1.20}}
% & \score{64.40}{\down{-1.03}}
\\

AdaBlock-dLLM
& \score{76.88}{\down{-0.45}}
& \score{\textbf{75.06}}{\up{+0.46}}
& \score{\underline{71.95}}{\up{+0.23}}
& \score{75.74}{\down{-0.61}}
& \score{74.83}{\down{-0.15}}
& \score{\textbf{76.95}}{\up{+0.07}}
& \score{\underline{37.00}}{\down{-0.60}}
& \score{34.40}{\up{+0.40}}
% & \score{65.35}{\down{-0.08}}
\\

AdaBlock-Fast
& \score{76.65}{\down{-0.68}}
& \score{73.92}{\down{-0.68}}
& \score{71.27}{\down{-0.45}}
& \score{75.66}{\down{-0.69}}
& \score{75.21}{\up{+0.23}}
& \score{74.15}{\down{-2.73}}
& \score{34.20}{\down{-3.40}}
& \score{\textbf{36.20}}{\up{+2.20}}
% & \score{64.66}{\down{-0.77}}
\\

dKV-Cache-Decode
& \score{72.93}{\down{-4.40}}
& \score{69.45}{\down{-5.15}}
& \score{66.94}{\down{-4.78}}
& \score{72.93}{\down{-3.42}}
& \score{72.63}{\down{-2.35}}
& \score{73.92}{\down{-2.96}}
& \score{31.80}{\down{-5.80}}
& \score{33.00}{\down{-1.00}}
% & \score{61.70}{\down{-3.73}}
\\

Elastic-Cache
& \score{\underline{77.10}}{\down{-0.23}}
& \score{73.84}{\down{-0.76}}
& \score{\textbf{72.18}}{\up{+0.46}}
& \score{\underline{75.97}}{\down{-0.38}}
& \score{\textbf{76.88}}{\up{+1.90}}
& \score{76.35}{\down{-0.53}}
& \score{33.60}{\down{-4.00}}
& \score{31.80}{\down{-2.20}}
% & \score{64.72}{\down{-0.71}}
\\

WINO
& \score{74.68}{\down{-2.65}}
& \score{72.48}{\down{-2.12}}
& \score{70.74}{\down{-0.98}}
& \score{74.45}{\down{-1.90}}
& \score{72.78}{\down{-2.20}}
& \score{75.21}{\down{-1.67}}
& \score{33.40}{\down{-4.00}}
& \score{34.20}{\up{+0.20}}
% & \score{63.49}{\down{-1.94}}
\\

\bottomrule
\end{tabular}
}}
\end{table}

\paragraph{dLLM decoding methods exhibit pervasive evaluation inconsistency}
Table~\ref{tab:math_llada_results} and Table~\ref{tab:math_llada1.5_results} report the accuracy of various decoding methods on math datasets across various prompt templates under LLaDA-8B-Instruct~\citep{nie2026large} and LLaDA-1.5~\citep{zhullada}, respectively. 
The results reveal that the performance of decoding methods is highly sensitive to prompt template choices, exhibiting substantial instability across prompt templates. 
For instance, in Table~\ref{tab:math_llada_results}, Elastic-Cache achieves better performance than Vanilla under Template 3, but Vanilla substantially outperforms Elastic-Cache under Template 2. 
We also report evaluation results for the Dream model in Appendix~\ref{dream_math}, where we observe a similar pattern.
Figure~\ref{Kendall's_llada8b} and Figure~\ref{Kendall's_llada1.5} in Appendix~\ref{ranking} show the pairwise Kendall's tau correlations (see definition in Appendix~\ref{Kendall}) among method rankings obtained under different prompt templates for LLaDA-8B-Instruct and LLaDA-1.5, respectively.
Higher values indicate more consistent relative ranking of decoding methods, whereas lower values reflect greater disagreement in method ranking across prompt templates. 
The average pairwise Kendall's tau across prompt templates is only 0.539 for LLaDA-8B-Instruct and 0.477 for LLaDA-1.5, indicating pervasive evaluation inconsistency across dLLM decoding methods.
In Appendix~\ref{tab:bootstrap}, we report bootstrap confidence intervals, confirming that the observed evaluation inconsistency is not an artifact of a specific test sample.
\textbf{These findings suggest that evaluating decoding methods under a single prompt template may undermine the reliability of evaluations.}
% \input{table/methods_rank2}

% \begin{figure*}[t!]
%     \centering
%     \begin{subfigure}[t]{0.45\textwidth}
%         \centering
%         \includegraphics[width=\linewidth]{fig/Kendall's tau1.pdf}
%         \caption{LLaDA-8B-Instruct}
%         \label{Kendall's_llada8b}
%     \end{subfigure}
%     \hspace{0.04cm} % 调整左右间距
%     \begin{subfigure}[t]{0.45\textwidth}
%         \centering
%         \includegraphics[width=\linewidth]{fig/Kendall's tau2.pdf}
%         \caption{LLaDA-1.5}
%         \label{Kendall's_llada1.5}
%     \end{subfigure}
%     \caption{Pairwise Kendall's tau correlation matrices across prompt templates on GSM8K for (a) LLaDA-8B-Instruct and (b) LLaDA-1.5. Each cell quantifies the ranking agreement of decoding methods between two prompt templates, with higher values indicating more consistent rankings.}
% \end{figure*}

% Through comprehensive experiments across prompt templates, we find that the performance gains of parallel decoding methods are highly prompt-dependent and that aggressive acceleration generation can incur accuracy degradation on downstream tasks.

\begin{figure*}[t!]
    % \vspace{-10mm}
    \centering
    \begin{subfigure}[t]{0.49\textwidth}
        \centering
        \includegraphics[
            width=\linewidth,
            height=0.23\textheight
        ]{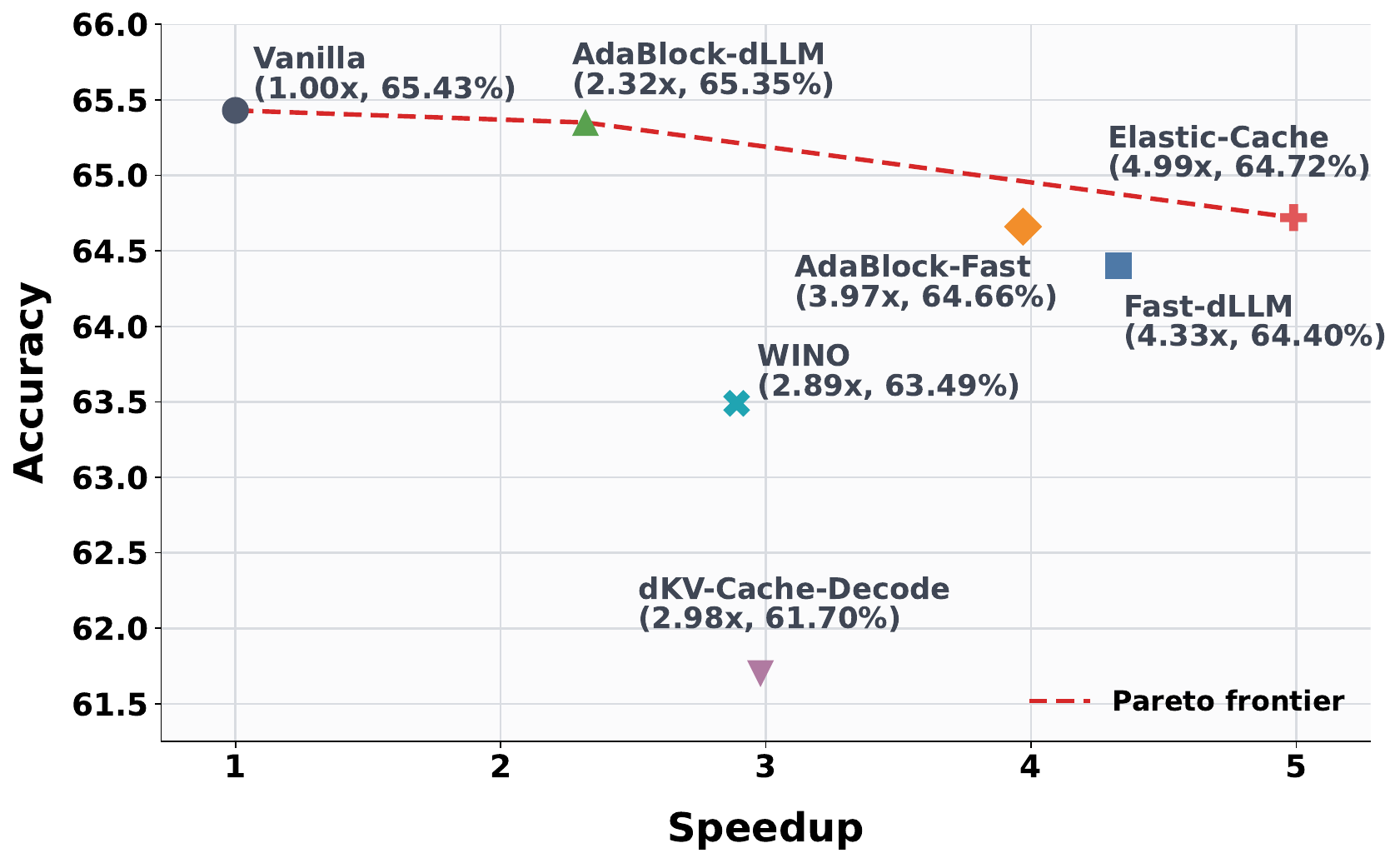}
        % \caption{Accuracy across varing thresholds $\gamma$}
        \caption{LLaDA-8B-Instruct}
        \label{fig:tradeoff_1}
    \end{subfigure}
    \hspace{0.1cm} %
    \begin{subfigure}[t]{0.49\textwidth}
        \centering
        \includegraphics[
            width=\linewidth,
            height=0.23\textheight
        ]{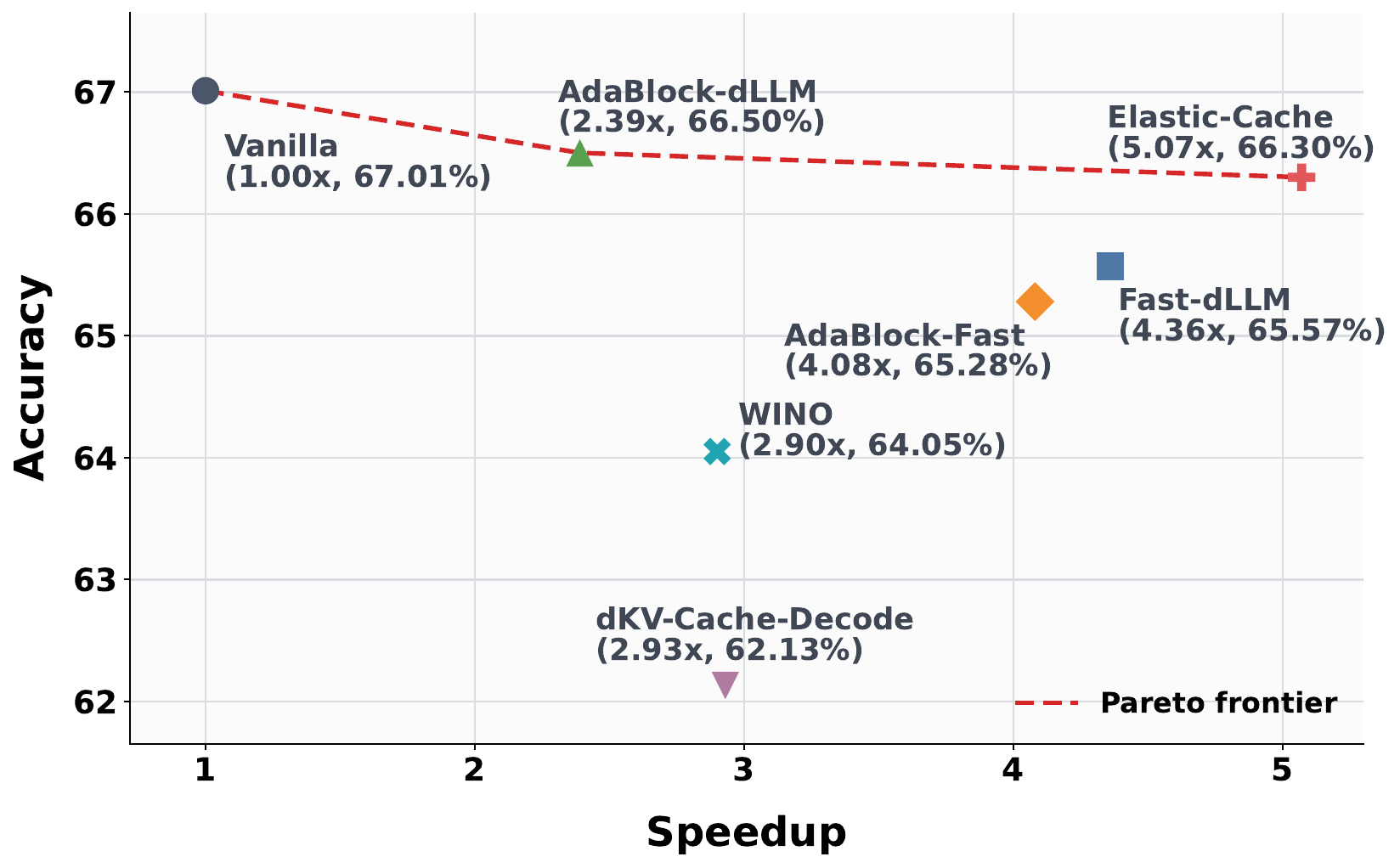}
        % \caption{Accuracy across varying prompts}
        \caption{LLaDA-1.5}
        \label{fig:tradeoff_2}
    \end{subfigure}
 \caption{The accuracy--speedup trade-offs of different decoding methods for (a) LLaDA-8B-Instruct and (b) LLaDA-1.5. The x-axis shows the average decoding speedup relative to Vanilla, and the y-axis shows the average accuracy across prompt templates.}
    % \label{fig:prompt}
    \vspace{-3mm}
\end{figure*}

\paragraph{Performance gains of decoding methods are highly prompt-dependent}
A fundamental challenge in dLLM decoding methods is the dilemma between inference efficiency and generation quality.
To address this, recent studies have explored parallel decoding methods designed to accelerate generation without sacrificing generation quality.
Table~\ref{tab:math_llada_results} and Table~\ref{tab:math_llada1.5_results} compare the accuracy of the vanilla baseline and current parallel decoding methods across various prompt templates.
Although some decoding methods occasionally produce marginal \textcolor{mydarkgreen}{improvements} over the vanilla baseline on certain prompt templates, \textcolor{mydarkred}{degradation} is widespread across the majority of prompt templates.
The evaluation results on code benchmarks in Appendix~\ref{code_benchmark} also show a similar trend.
These results show that the performance gains of parallel decoding methods are highly prompt-dependent and that aggressively accelerating generation can incur accuracy degradation on downstream tasks.
More detailed results are provided in Appendix~\ref{trade-off}.
We further present the accuracy–speedup trade-off of decoding methods in Figure~\ref{fig:tradeoff_1} and Figure~\ref{fig:tradeoff_2}. 
The x-axis shows the average decoding speedup relative to Vanilla, computed from the measured end-to-end decoding times~\citep{wu2026fastdllm}, while the y-axis shows the average accuracy across prompt templates. 
We find that Vanilla, AdaBlock-dLLM, and Elastic-Cache form the Pareto frontier across both models, achieving the superior trade-offs between accuracy and decoding speed.
% In particular, the average accuracy across prompt templates further suggests that current parallel decoding methods accelerate generation at a non-trivial cost to accuracy on downstream tasks.
% Additionally, we find that methods without KV caching (e.g., AdaBlock-dLLM) outperform methods employing KV caching (e.g., Fast-dLLM ), suggesting that KV caching can result in performance degradation.
\textbf{Our findings suggest that evaluating dLLM decoding methods relying on a single prompt template can undermine the reliability of evaluation.}

% \textbf{These results suggest that relying on a certain prompt template risks creating an evaluation illusion that current parallel decoding methods can simultaneously accelerate generation and maintain generation quality.}

% \paragraph{KV cache-based decoding methods incur performance degradation}
% \section{Understanding Performance Dynamics of dLLM Decoding Methods}
\section{Performance Dynamics of dLLM Decoding Methods}
% Autoregressive models generate text in a left-to-right order, while dLLMs iteratively determine both what tokens to generate and where to unmask at each decoding step.
% This flexibility makes the decoding trajectory jointly dependent on token prediction and position selection. 
% To uncover the mechanism driving prompt sensitivity in dLLM decoding, we conducted systematic experiments to investigate how prompts affect the dynamics of dLLM generation. 
% \subsection{Investigating Sensitivity of Decoding Methods to Minor Prompt Variations}
% \subsection{Investigating the Sensitivity of dLLM Decoding Methods to Minor Prompt Variations}
\subsection{Sensitivity of Decoding Methods to Minor Prompt Variations}
% \paragraph{Experimental setup}
% Based on the previous observation that decoding methods are highly sensitive to commonly used prompt templates, we are motivated to systematically examine how minor prompt variations affect performance metrics in evaluation.
% Specifically, we construct a set of near-identical prompt templates that differ only in individual words or characters, as presented in Appendix~\ref{label:prompt_subtle}.
% We evaluate three decoding strategies on GSM8K under LLaDA-8B-Instruct: single-token decoding (Vanilla), Top-K decoding, and threshold-based confidence decoding.
% We provide implementation details in Appendix~\ref{implementation}.
% Vanilla decoding employs low-confidence remasking, committing only the highest-confidence masked token at each step, and generally achieves superior performance.
% Parallel decoding accelerates generation by decoding multiple tokens per step, reducing denoising steps, but can degrade task performance.
% The Top-K decoding method selects the $k$ masked tokens with the highest confidence scores, while the threshold-based confidence decoding method unmasks all tokens whose confidence exceeds a predefined threshold.

\paragraph{Experimental setup}
Motivated by the high sensitivity of decoding methods to prompt templates, we systematically examine how minor prompt variations affect evaluation metrics.
Specifically, we construct near-identical prompt templates differing only in individual words or characters (Appendix~\ref{label:prompt_subtle}).
We evaluate three decoding strategies on GSM8K under LLaDA-8B-Instruct: single-token decoding (Vanilla), Top-K decoding, and threshold-based confidence decoding.
We provide implementation details in Appendix~\ref{implementation}.
Vanilla decoding employs low-confidence remasking, committing only the highest-confidence masked token at each step, and generally achieves superior performance.
Parallel decoding accelerates generation by decoding multiple tokens per step, reducing denoising steps, but can degrade task performance.
The Top-K decoding method selects the $k$ masked tokens with the highest confidence scores, while the threshold-based confidence decoding method unmasks all tokens whose confidence exceeds a predefined threshold.

% \subsection{Understanding the Mechanism of Prompt Sensitivity in Parallel Decoding}

\begin{figure*}[t!]
    % \vspace{-10mm}
    \centering
        \begin{subfigure}[t]{0.325\textwidth}
        \centering
        \includegraphics[width=\linewidth]{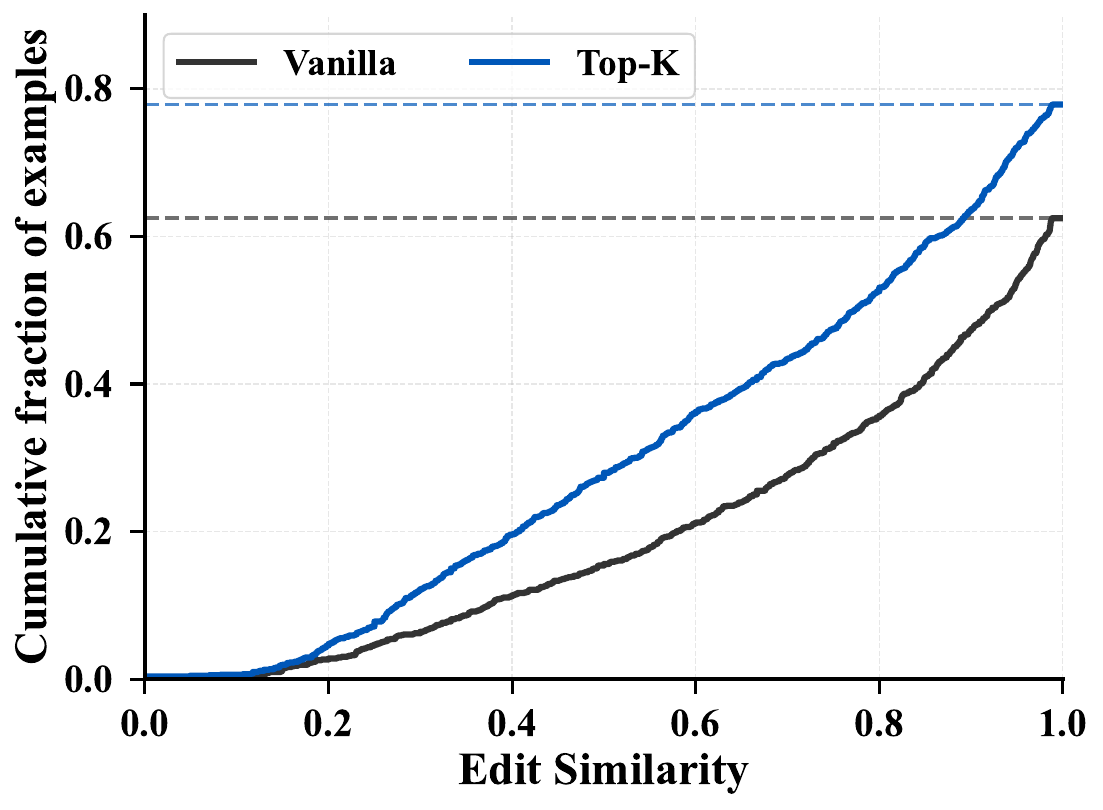}
        \caption{Distribution of edit similarity}
        \label{fig:edit_simi_two}
    \end{subfigure}
    \begin{subfigure}[t]{0.325\textwidth}
        \centering
        \includegraphics[width=\linewidth]{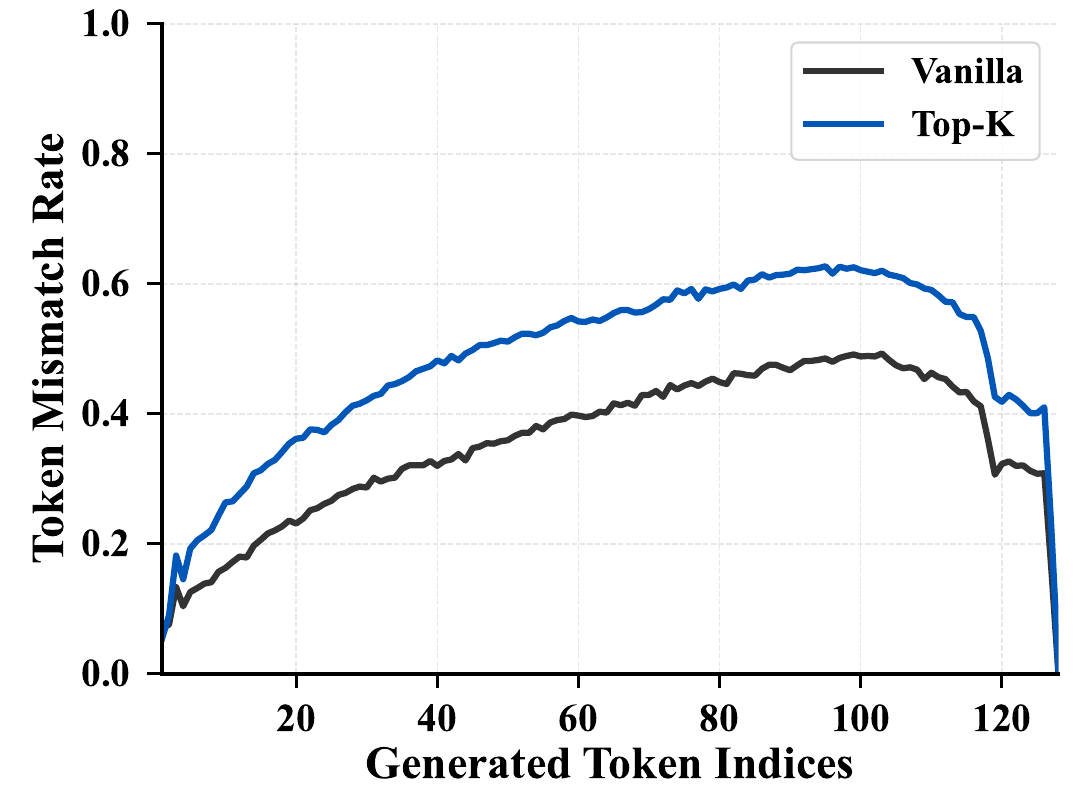}
        % \caption{Accuracy across varying prompts}
        \caption{Token mismatch rate}
        \label{fig:mismatch}
    \end{subfigure}
    \begin{subfigure}[t]{0.325\textwidth}
        \centering
        \includegraphics[width=\linewidth]{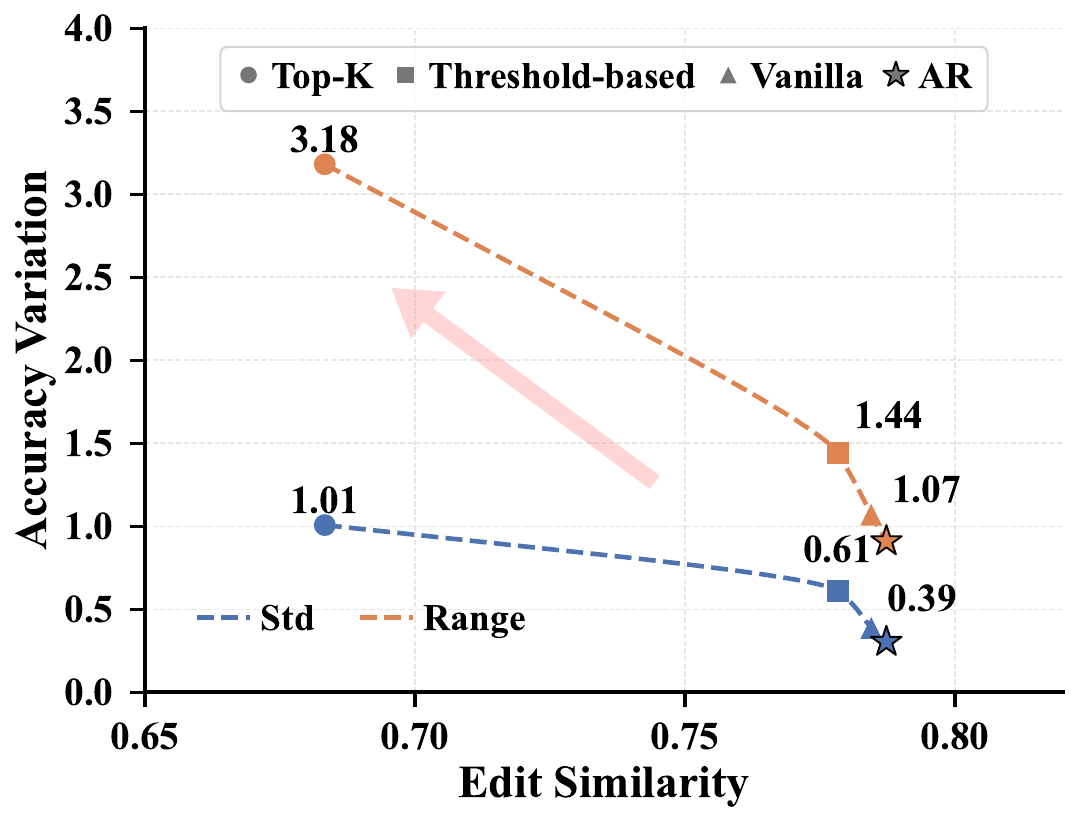}
        \caption{Similarity vs. accuracy variation}
        \label{fig:edit_simi}
    \end{subfigure}
    
\caption{
Prompt sensitivity of parallel decoding methods under near-identical prompt templates.
(a) Cumulative fraction of examples as a function of edit similarity under prompt variations for Vanilla and Top-K decoding.
(b) Position-wise token mismatch rates for Vanilla and Top-K decoding across prompt pairs.
(c) Edit similarity vs. accuracy variation across AR and dLLM decoding methods, including Vanilla, threshold-based, and Top-K.
}
\vspace{-5mm}
\end{figure*}

\paragraph{Parallel decoding transfers subtle prompt variations into inconsistent outputs}
We analyze how minor prompt variations affect dLLM generation behavior by measuring output similarity under pairs of near-identical prompts, as presented in Example~\ref{box:discrepancies}.
In our experiment, we use Top-K (k=2) decoding as a parallel decoding baseline, compared with single-token decoding baselines.
We plot the cumulative distribution function (CDF) of edit similarities between generated outputs under the two prompt templates in Figure~\ref{fig:edit_simi_two}.
As defined in Appendix~\ref{edit_similarity}, we employ edit similarity to quantify token-level agreement between generated sequences, where higher scores reflect greater output consistency across prompt templates.
The higher CDF curve for Top-K decoding reflects a higher proportion of low-similarity outputs across prompts, suggesting that Top-K decoding increases token-level divergence under subtle prompt variations. 
We further examine the position-wise token mismatch across prompt pairs in Figure~\ref{fig:mismatch}.
The results show that Top-K exhibits a higher token mismatch rate at early output positions, which subsequently propagates and amplifies across later generation positions. 
Ultimately, the mismatch rate drops sharply toward the end of the sequence due to \textit{EOS} token padding.
These results indicate that parallel decoding is more prone to produce inconsistent outputs under subtle prompt variations. 
In practice, AR models generate text in a left-to-right order, while dLLMs iteratively determine both what tokens to generate and where to unmask at each decoding step.
This flexibility makes the decoding trajectory jointly dependent on token prediction and position selection. 
In Appendix~\ref{divergence}, we present a case study illustrating how subtle prompt variations steer generation dynamics toward divergent trajectories.

\paragraph{Output divergence induces performance variations across prompts}
To examine how output divergence affects accuracy variation, we evaluate output consistency and accuracy variation across near-identical prompts (see Appendix~\ref{label:prompt_subtle}) for vanilla single-token decoding and for parallel decoding strategies.
Figure~\ref{fig:edit_simi} reveals a clear pattern between edit similarity and accuracy variation, where lower output similarity corresponds to larger performance fluctuations across prompt templates.
Consequently, parallel decoding amplifies sensitivity to subtle prompt perturbations by heightening output divergence.
In Appendix~\ref{divergence}, we provide a case study indicating that subtle prompt variations can induce divergent reasoning trajectories and yield different final answers.
We further evaluate the autoregressive model using Qwen3-8B~\citep{yang2025qwen3technicalreport} across various prompt templates.
The AR baseline exhibits substantially lower variation in accuracy than dLLMs, as the next-token prediction paradigm inherently offers less decoding flexibility.
These findings indicate that the greater decoding freedom in dLLMs promotes divergent outputs, resulting in greater variation in task performance.

\paragraph{Parallel decoding is more susceptible to minor prompt variations}
Table~\ref{tab:prompt} in Appendix~\ref{minior_prompt} presents the accuracy of the single-token decoding method and different parallel decoding strategies across various near-identical prompt templates on GSM8K under LLaDA-8B-Instruct. 
We quantify the sensitivity of decoding methods to prompt variation using the standard deviation and range of accuracy across prompt templates.
% Surprisingly, even minor variations in prompt templates can substantially affect the performance of the decoding methods evaluated.
Our results show that parallel decoding methods exhibit substantially greater sensitivity to minor variations of the prompt template.
In practice, decoding more tokens per step leads to greater performance variability, as more concurrently decoded tokens are more prone to inter-token dependency conflicts.
Moreover, Top-K methods exhibit considerably higher standard deviation and range than threshold-based decoding methods, suggesting that they are more susceptible to prompt template variations.
This stems from Top-K decoding potentially sampling low-confidence tokens, which disrupts token dependencies under the conditional independence assumption, thereby amplifying sensitivity to minor prompt variations.
\textbf{These findings indicate that parallel decoding methods are particularly susceptible to inconsistent evaluation results due to their high sensitivity to prompt templates.}

% \subsection{How minor prompt variations affect the performance of decoding methods}
% Comprared wth AR models

\subsection{Effects of Decoding Budget and Prompt Templates on Evaluation}
\paragraph{Experimental setup}
Diffusion large language models generate text through iterative denoising, progressively unmasking tokens across multiple denoising steps. 
While increasing the number of denoising steps is generally expected to improve generation quality, a natural question arises: \textit{Do more denoising steps necessarily lead to better performance on downstream tasks?}
In this subsection, we empirically investigate how the number of denoising steps and the choice of prompt template jointly affect evaluation performance.
In our experiments, we construct a set of semantically equivalent prompt templates A--H with lexical and stylistic variations but identical semantics and output format instructions, as presented in Appendix~\ref{label:prompts_step}. 
We conduct experiments with threshold-based parallel decoding across varying confidence thresholds $\gamma \in \{0.6, 0.7, 0.8, 0.9 \}$, obtaining varying numbers of denoising steps per generation. 
The detailed results of accuracy and numbers of denoising steps of each decoding method are presented in Appendix~\ref{label:acc_step}.

\begin{figure*}[t!]
    \centering
    % \vspace{-10mm}
    \begin{subfigure}[t]{0.46\textwidth}
        \centering
        \includegraphics[width=\linewidth]{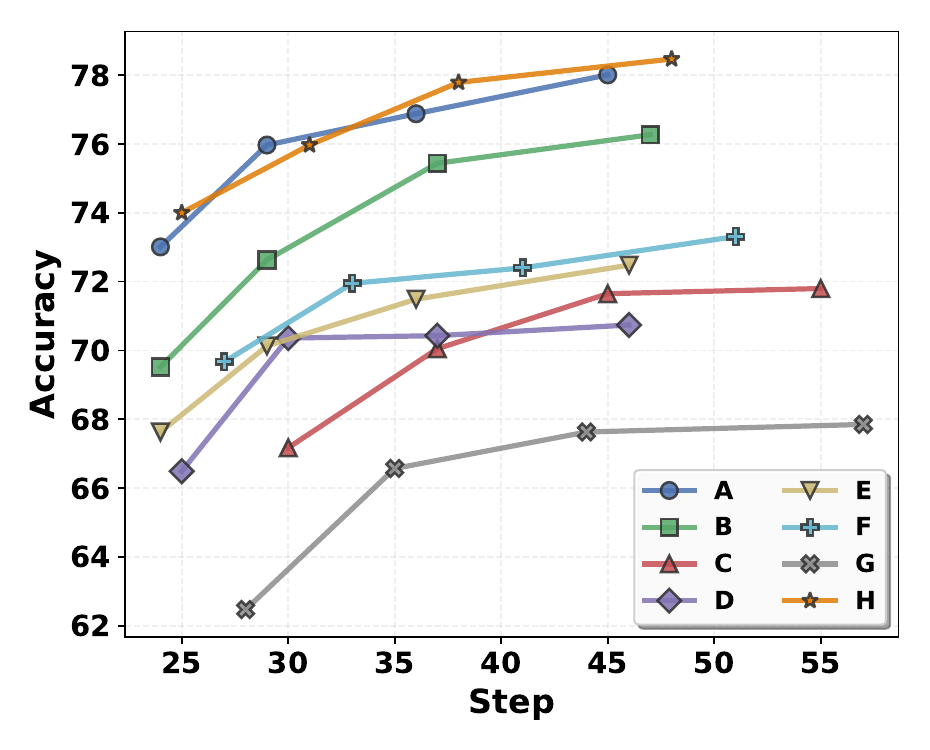}
        % \caption{Accuracy across varing thresholds $\gamma$}
        \caption{Accuracy vs. decoding steps within prompts}
        \label{fig:acc_step}
    \end{subfigure}
    \hspace{0.04cm} %
    \begin{subfigure}[t]{0.46\textwidth}
        \centering
        \includegraphics[width=\linewidth]{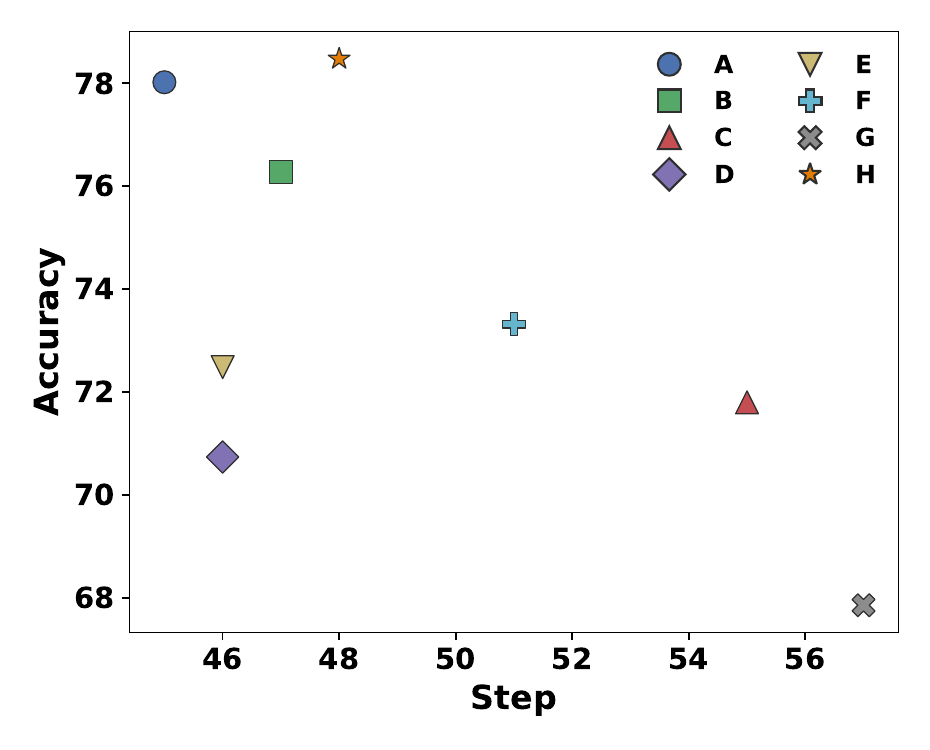}
        % \caption{Accuracy across varying prompts}
        \caption{Accuracy vs. decoding steps across prompts}
        \label{fig:acc_prompts}
    \end{subfigure}
    \caption{Accuracy of parallel decoding method across a set of semantically equivalent prompt templates. (a) Accuracy of threshold-based parallel decoding method with varying thresholds $\gamma \in \{0.6, 0.7, 0.8, 0.9 \}$ across various prompts. (b) Accuracy of threshold-based parallel decoding method ($\gamma=0.9$) across various prompts.}
    \vspace{-3mm}
\end{figure*}

\paragraph{Increasing decoding budget improves accuracy within each prompt template}
Figure~\ref{fig:acc_step} reports the accuracy of the threshold-based parallel decoding method across various prompt templates on GSM8K under LLaDA-8B-Instruct, where different confidence thresholds $\gamma \in \{0.6, 0.7, 0.8, 0.9 \}$ produce varying numbers of denoising steps. 
The results show that increasing the number of denoising steps consistently improves accuracy within each prompt template. \textit{This suggests an inherent trade-off between decoding budget and accuracy, motivating future methods that can overcome this limitation.}

\paragraph{A larger decoding budget does not necessarily yield better performance across prompts}
Figure~\ref{fig:acc_prompts} shows the accuracy of threshold-based decoding ($\gamma$=0.9) across different prompt templates (A-H), where each point corresponds to a prompt template, with the x-axis indicating the average number of denoising steps and the y-axis indicating the accuracy. 
Notably, even when the number of denoising steps is comparable, accuracy can vary substantially across prompt templates.
For instance, template H achieves 78.47\% accuracy with 48 steps, while template G achieves only 67.85\% accuracy despite consuming 57 steps, a significant degradation of 10.62 percentage points. 
\textbf{This reveals that performance differences cannot be attributed solely to decoding budget, as a well-designed prompt template can achieve superior performance with fewer steps.}

\begin{wrapfigure}{r}{0.45\textwidth}
    \centering
    % \vspace{-5mm}
    \includegraphics[width=\linewidth]{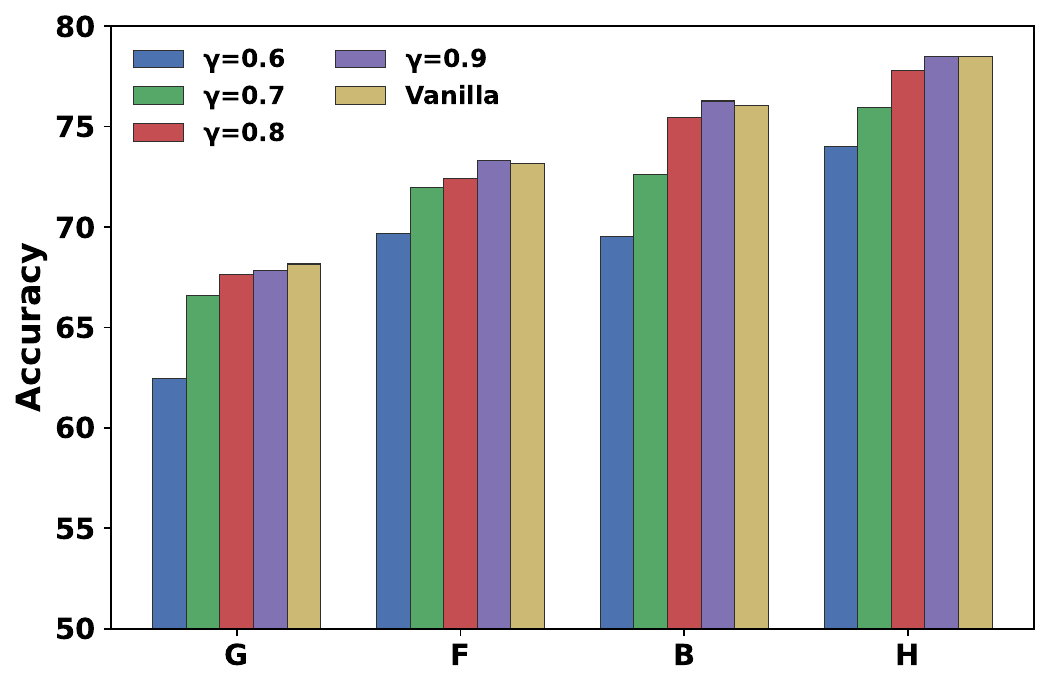}
    \caption{Accuracy variability of various decoding methods across a set of semantically equivalent prompt templates.}
    \label{fig:prompt_adv}
    \vspace{-4mm}
\end{wrapfigure}

\paragraph{Prompt templates dominate evaluation results}
Figure~\ref{fig:prompt_adv} shows the accuracy of threshold-based decoding under varying confidence thresholds, and the Vanilla baseline across four semantically equivalent prompt templates (i.e., G, F, B, H) on GSM8K under LLaDA-8B-Instruct.
The experimental results show that prompt template choice has a dominant effect on evaluation metrics. 
Surprisingly, while increasing the confidence threshold yields moderate performance gains within a given prompt template, the choice of prompt template has a far greater impact on performance.
For instance, parallel decoding under prompt template H with a low acceptance threshold of 0.6 achieves 74\% accuracy within only 25 steps, substantially outperforming vanilla decoding under prompt template G, which demands 128 denoising steps while yielding a remarkably lower accuracy of 68.16\%.
An effective prompt template can consistently achieve superior performance across decoding configurations.
These findings indicate that reported performance gains in decoding methods may stem from the choice of prompt template, underscoring the need for multi-prompt evaluation.

% \textbf{A favorable prompt template consistently yields superior accuracy across all 
% decoding configurations, suggesting that performance gains observed in the evaluation 
% of decoding strategies may stem from prompt selection rather than their authentic effectiveness.}

\subsection{Potential Factors Contributing to Variation in Evaluation Metrics}
\paragraph{How does few-shot evaluation affect the evaluation of decoding methods?}
To investigate the effect of few-shot prompting on the performance of decoding methods, we conduct experiments with current decoding methods under different few-shot prompting settings. 
Table~\ref{tab:few_shot} in Appendix~\ref{few_shots} reports the accuracy of vanilla and parallel decoding methods across varying few-shot settings on GSM8K under LLaDA-8B-Instruct.
Few-shot evaluation consistently improves accuracy across all methods, yet the performance gain differs substantially between single token decoding and parallel decoding methods. 
The Vanilla baseline achieves superior performance with 2-shot examples, followed by a decline.
In contrast, parallel decoding methods achieve their best performance in settings with more examples of few shots, such as the 4-shot setting.
The experimental results indicate that varying the number of few-shot examples markedly affects the evaluation results of dLLM decoding methods.
Furthermore, we observe that KV-reuse decoding methods like Fast-dLLM exhibit a larger inference speedup advantage as the number of few-shot examples grows. 
This is because longer prompts increase the overhead of repeatedly encoding the prefix across denoising steps, whereas KV reuse largely eliminates this redundant computation.
Therefore, the choice of a few-shot configuration dictates not only accuracy but also the comparative inference efficiency of decoding methods.
Overall, the few-shot configuration substantially influences both the observed accuracy and the relative inference efficiency between the decoding methods.

% This suggests that the number of few-shot examples can affect the evaluation results of dLLM decoding methods.

% \subsection{data type}
% Furthermore, parallel decoding is more sensitive to hardware platforms than single-token decoding, with higher average and maximum deviations in accuracy across all prompt templates. 
% These findings suggest that accuracy variation across GPU types is especially pronounced in parallel decoding methods, where simultaneous unmasking of multiple tokens per step amplifies floating-point rounding error~\citep{yuan2026understanding}, ultimately leading to greater accuracy variation.

\begin{wrapfigure}{r}{0.5\textwidth}
    \vspace{-4mm}
    \centering
    \includegraphics[width=\linewidth]{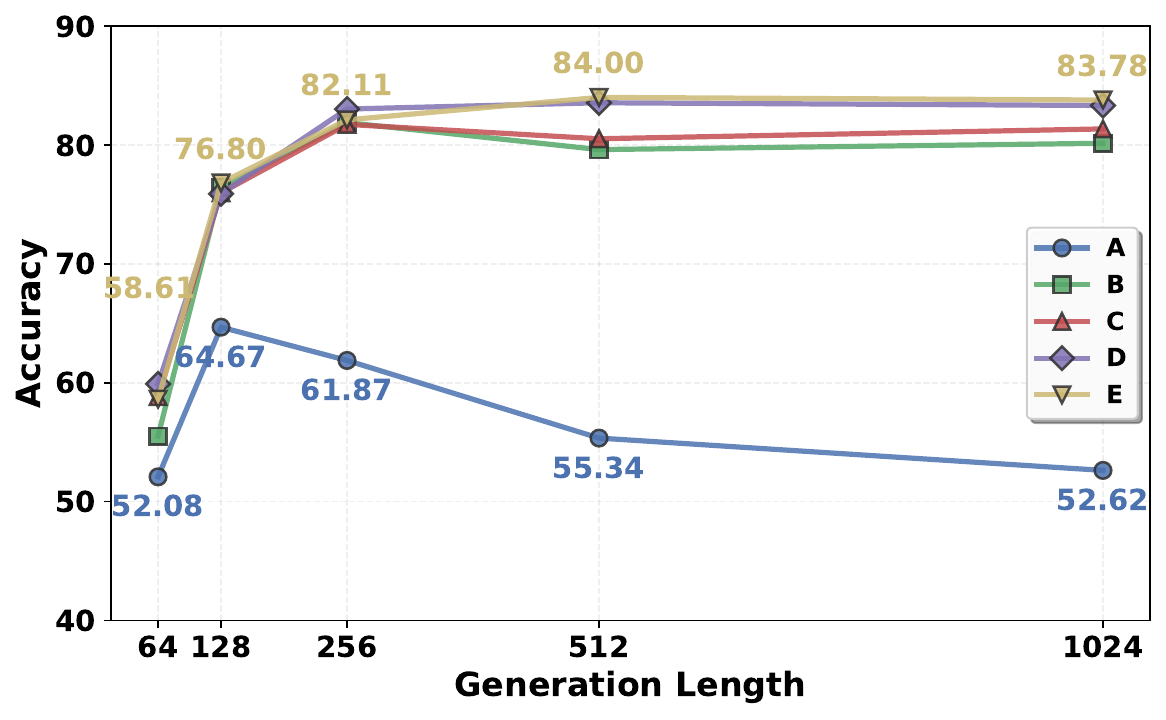}
    \caption{Accuracy of threshold-based parallel decoding under different generation lengths across prompt templates.}
    \label{fig:acc_cot}
    \vspace{-4mm}
\end{wrapfigure}
\paragraph{How does generation length affect evaluation results?}
dLLMs generate text by iteratively denoising over a fixed generation length, and increasing the generation length is generally expected to improve generation quality.
However, our experiments reveal that increasing generation length does not consistently yield performance gains. 
Specifically, we construct a set of zero-shot CoT prompt templates with identical task instructions and output format requirements, as presented in Appendix~\ref{label:prompt_cot}.
These prompt templates differ only in their reasoning instructions, ranging from no explicit reasoning instructions (Prompt A) to increasingly detailed ones (Prompts B–E).
Figure~\ref{fig:acc_cot} shows the accuracy of the threshold-based parallel decoding method ($\gamma = 0.8$) with varying generation length on GSM8K.
The results show that accuracy improves as the generation length increases from 64 to 128 across all prompt templates, suggesting that an insufficient generation length may constrain the model's reasoning process and thereby harm accuracy.
For prompt templates B–E, accuracy consistently improves with increasing generation length, stabilizing beyond 512.
In contrast, for prompt template A, accuracy degrades when the generation length exceeds 128, as a larger generation length may create a mismatch between the predefined length and the task-required response length.
We find that under template A, the model tends to skip intermediate reasoning steps and directly output final answers when the generation length exceeds 128, degrading generation quality.
Our findings show that different prompt templates exhibit inconsistent preferences for generation length, with increasing generation length sometimes degrading performance for certain prompts. 

\paragraph{Conducting evaluations across GPU types risks unreliable evaluation results.}
Table~\ref{tab:prompt_gpu} in Appendix~\ref{label:float} reports the accuracy discrepancies across different GPU types, including L40, Pro6000, A100, and RTX 4090, under identical decoding method and evaluation settings. 
We observe that non-trivial accuracy discrepancies arise across GPU types even under identical experimental conditions, indicating that GPU types are a non-negligible factor in evaluation.
This performance difference across GPU types stems from the non-associative nature of floating-point arithmetic under limited numerical precision~\citep{yuan2026understanding}.
Our analysis shows that evaluating decoding methods across different hardware platforms can introduce evaluation inconsistency, potentially affecting the reliability of conclusions.
Furthermore, we also conduct experiments across GPU types under FP32 precision. 
Table~\ref{tab:gpu_float32} in Appendix~\ref{label:float} reports the accuracy discrepancies across GPU types under identical decoding methods and prompt templates, evaluated in FP32 precision.
The results show that the evaluation results on L40 and A100 are nearly identical, suggesting that adopting FP32 precision can substantially mitigate the evaluation variance introduced by hardware.

% This may account for why evaluation in recent works with a larger generation length can yield lower benchmark scores than those with a shorter one.

\section{Discussion}
% Prior work regarding decoding methods for dLLMs has made substantial efforts to advance our understanding of this field. 
Prior work on decoding methods for dLLMs has primarily focused on improving inference efficiency, with little attention paid to the reliability of their evaluation.
By revealing evaluation concerns in our community, our goal is to establish more reliable evaluation practices that advance future research.
Based on our findings, we propose the following practical suggestions for evaluating decoding methods in diffusion large language models.
\begin{itemize}
    \item Report performance metrics across multiple prompt templates. Evaluating dLLM decoding methods with a single prompt template can undermine the reliability of the evaluations. Our experiments in Appendix~\ref{multi_evaluation} show that multi-prompt evaluation can effectively mitigate this evaluation inconsistency.
    
    \item Conduct evaluations on identical GPU platforms. When conducting evaluations across different GPU platforms, we recommend adopting FP32 precision or other more efficient techniques for deterministic inference, such as LayerCast~\citep{yuan2026understanding}, to mitigate evaluation variance introduced by hardware platforms.
    \item Provide detailed evaluation settings. We recommend thoroughly reporting the evaluation configurations, including the employed prompt template, algorithmic parameters, generation configurations (e.g., generation length), and hardware specifications (e.g., GPU type) to ensure reproducibility.
\end{itemize}

\section{Conclusion}
In this work, we conduct a rigorous evaluation of current decoding methods for dLLMs across evaluation settings.
We find that the performance of parallel decoding methods is highly prompt-dependent, undermining the reliability of single-prompt evaluation protocols.
Through comprehensive experiments, we attribute this evaluation inconsistency to the high sensitivity of parallel decoding methods to subtle prompt template variations. 
These findings highlight the importance of multi-prompt evaluation for dLLM decoding methods.
Furthermore, our experiments indicate that other overlooked evaluation settings (e.g., hardware) can also notably affect the assessment of decoding methods.
Overall, our work aims to provide guidelines for uncovering potential evaluation discrepancies and facilitating fair, reproducible evaluations.
We hope that our study deepens the understanding of the evaluation challenges in dLLMs and advances future research.
% \paragraph{Limitations}
% Our work aims to reveal potential evaluation issues for evaluating dLLM decoding methods.
% Although we provide practical guidelines to promote reliable evaluation,  a fully standardized, comprehensive evaluation framework remains an important direction for future work.

% We find that prompt templates dominate evaluation results, suggesting that the performance gains of decoding methods may stem from the choice of prompt template.
% Additionally, our experiments indicate that other overlooked evaluation settings (e.g., hardware) can also notably affect the assessment of decoding methods.
% Ultimately, we highlight several practical considerations for more reliable evaluation of decoding methods in dLLMs, such as conducting multi-prompt evaluation and transparently reporting evaluation settings.

\clearpage

% \section*{Ethics Statement}
% This work aims to improve the reliability and reproducibility of evaluations for
% diffusion large language models. 
% All experiments are conducted on public benchmarks and involve no human subjects, private data, or personally identifiable information. 
% We identify no direct societal risks beyond those broadly associated with language model research. 
% Our work conducts evaluation using existing open-source models and releases no new models, posing no additional safety risks.
% This work aims to shed light on evaluation illusions in dLLMs and guide future research toward
% more reliable assessment practices.

% \section*{Reproducibility Statement}
% To ensure reproducibility, we provide detailed implementation parameters in Appendix~\ref{details}. 
% Additionally, we release experimental code in an~\href{https://anonymous.4open.science/r/evaluation_dllms-96D2}{anonymous repository}. 
% We will release the evaluation code to facilitate reproducibility and future research. 

% \section*{Use of Large Language Models}
% In our work, large language models were used solely for literature search and language polishing.
% The authors are responsible for all technical contributions, experimental implementations, and evaluation results.

\bibliography{iclr2027_conference}
\bibliographystyle{iclr2027_conference}

\appendix
\clearpage

\addcontentsline{toc}{section}{Appendix}
\part{Appendix} % Start the appendix part
\parttoc % Insert the appendix TOC

% \section{Impact Statement and Limitations}
% \label{impact}
% \paragraph{Impact Statement} 
% This work aims to improve the reliability and reproducibility of evaluations for diffusion large language models.
% All experiments are conducted on public benchmarks and involve no human subjects, private data, or personal identification information. 
% We identify no direct societal risks beyond those broadly associated with language model research.
% Our work conducts evaluation using existing open-source models and releases no new models, posing no additional safety risks.
% This work aims to shed light on evaluation illusions in dLLMs and guide future research toward more reliable assessment practices.

% \paragraph{Limitations}
% Our work reveals evaluation concerns in dLLMs: the ranking of decoding methods can vary significantly across prompt templates, undermining the reliability of evaluations under a single prompt template.
% Although we provide practical guidelines for reliable evaluation of decoding methods in dLLMs, a more systematic evaluation framework remains an important direction for future work.

\section{Related Work}
\paragraph{Decoding methods for dLLMs}
Recent efforts to accelerate dLLM inference have focused on two primary challenges: reducing the computational cost per denoising step and decreasing the number of denoising steps~\citep{hong2026widein,li2026diffusion}.
The bidirectional attention of dLLMs precludes the use of standard autoregressive KV caching, motivating several works to propose approximate caching strategies.
dKV-Cache proposes delayed and conditioned KV caching, while Fast-dLLM introduces block-wise approximate KV caching tailored for bidirectional diffusion models~\citep{ma2025dkvcache,wu2026fastdllm}.
Other methods further reduce computational overhead per step by exploiting the stability of the token representation, applying selective cache refresh, or introducing sparse computation patterns~\citep{liu2025dllm,nguyen-tri2026attention,jiang2026dcache,song2026sparse,chen2026dpad}.
Additionally, some approaches aim to reduce the number of denoising steps by unmasking multiple tokens per step~\citep{wei2026accelerating,lu2026semanticaware,li2026diffusion,bao2026learning,wu2025free,agrawal2025spiffy,gao2025self}.
EB-Sampler uses an entropy-bounded unmasking strategy to control the number of decoded tokens per step~\citep{ben-hamu2026accelerated}.
WINO proposes a draft-and-verify decoding strategy that allows aggressively drafted tokens to be re-masked upon verification failure~\citep{hong2026widein}.
These methods have been reported to achieve substantial inference speedups without degradation in generation quality.
% Our work reveals an evaluation illusion in dLLMs: the ranking of decoding methods can vary significantly across prompt templates, undermining the reliability of evaluations with a single prompt template.
Our work reveals an evaluation illusion in dLLMs: method rankings vary significantly across prompt templates, undermining the reliability of single-prompt evaluations.

\paragraph{Evaluation robustness in LLMs}
LLM evaluation has been shown to be highly sensitive to seemingly minor design choices, such as prompt wording, formatting, and demonstration selection and ordering~\citep{jiang2020can,webson2022prompt,perez2021true,liu2022makes,zhao2021calibrate,lu2022fantastically}.
Prior work on in-context learning has shown that prompt format and demonstration order can cause substantial variance in few-shot accuracy~\citep{zhao2021calibrate,lu2022fantastically}.
Recent work also demonstrates that LLM performance is sensitive to formatting changes~\citep{sclar2024quantifying} and broader prompt formulation variations~\citep{razavi2025benchmarking}.
Adversarial perturbations at the character, word, sentence, and semantic levels further amplify this sensitivity~\citep{zhu2023promptrobust}.
To achieve reliable estimates, some work advocates evaluating models across multiple instruction templates and reporting performance scores aggregated over these templates~\citep{mizrahi2024state,polo2024efficient,sakai2024toward}.
Recent analysis further shows that some observed prompt sensitivity may stem from evaluation artifacts such as rigid answer matching or likelihood-based scoring~\citep{hua2025flaw}.
While these studies reveal the instability of LLM evaluation, they primarily focus on evaluating the performance of autoregressive LLMs.
By introducing an extra position selection dimension, dLLM decoding further amplifies prompt-induced generation instability compared to next-token prediction.
In our work, we investigate how prompt templates affect the evaluation of dLLM decoding methods, showing that parallel decoding methods are highly sensitive to the choice of prompt template.

\section{Experimental Details}
\label{details}
\subsection{Decoding methods for dLLMs}
\label{baselines}
Here, we summarize the decoding methods for diffusion language models examined in this work.
For each method, we use the best-performing hyperparameter settings reported in the corresponding paper or provided in the official codebase. 
We carefully verify these configurations during implementation and report all detailed hyperparameter settings below.

\paragraph{Low-confidence}
The low-confidence remasking strategy is a widely used decoding method for dLLMs~\citep{nie2026large}.
During generation, the model predicts all masked positions but commits only the most confident token, leaving the remaining positions masked for subsequent refinement.
The single-token decoding method achieves strong performance but requires a number of denoising steps equal to the generation length.
In experiments, we use this method as the \textit{Vanilla} baseline with generation length $L=128$ and block size $B=32$.

\paragraph{Fast-dLLM}
Fast-dLLM combines block-wise approximate KV caching with confidence-aware parallel decoding to accelerate inference in dLLMs~\citep{wu2026fastdllm}.
The caching component reuses approximate KV states across denoising steps within a block-wise decoding process, while the decoding component commits all predictions exceeding a confidence threshold and keeps the remaining positions masked.
Thus, Fast-dLLM reduces both redundant per-step computation and the number of denoising steps.
Following the official implementation of Fast-dLLM~\citep{wu2026fastdllm}, we evaluate Fast-dLLM with confidence threshold $\tau= 0.9$, and block size $B=32$.
For the Dream model, following the official codebase, we adopt the same hyperparameter settings as those used for LLaDA.

\paragraph{AdaBlock-dLLM}
AdaBlock-dLLM addresses the fixed block size limitation of semi-autoregressive dLLM decoding by adaptively determining the block size during generation~\citep{lu2026semanticaware}.
At the start of each block, the model inspects a local context window over the predicted sequence and examines the associated confidence scores. 
If a semantic delimiter is identified with high confidence above the threshold, the block boundary is placed at that position. 
Otherwise, the predefined default block size is adopted.
Following the official implementation of AdaBlock-dLLM~\citep{lu2026semanticaware}, we evaluate AdaBlock-dLLM with predefined default block size $B=32$, delimiter set $\mathcal{D}=[198]$, confidence threshold $\tau=0.9$ and delimiter threshold $\tau_D=0.3$.
For the Dream model, following the official codebase, we use the delimiter set $\mathcal{D}=\{198, 271, 280, 319, 340, 382, 401, 532, 624, 630, 692, 698,
921, 1248, 1837, 1939, 2219, 2533, 3276, \\3876, 4894, 5267, 14750, 68327\}$ and a delimiter threshold of $\tau_D=0.5$.

\paragraph{AdaBlock-Fast}
AdaBlock-Fast~\citep{lu2026semanticaware} integrates the adaptive block size determination mechanism of AdaBlock-dLLM into the Fast-dLLM decoding framework.
Following the official implementation of AdaBlock-dLLM~\citep{lu2026semanticaware}, we evaluate AdaBlock-Fast with predefined default block size $B=32$, delimiter set $\mathcal{D}=[198]$, confidence threshold $\tau=0.9$ and delimiter threshold $\tau_D=0.3$.
For the Dream model, following the official codebase, we use the delimiter set $\mathcal{D}=\{198, 271, 280, 319, 340, 382, 401, 532, 624, 630, 692, 698,
921, 1248, 1837, 1939, 2219, 2533, 3276, \\3876, 4894, 5267, 14750, 68327\}$ and a delimiter threshold of $\tau_D=0.5$.

\paragraph{dKV-Cache-Decode}
dKV-Cache adapts KV caching to diffusion language model decoding through a delayed and conditioned caching strategy~\citep{ma2025dkvcache}. 
The method exploits the observation that token representations stabilize at different rates during the denoising process. 
Instead of immediately reusing all KV states, it delays cache reuse and conditions caching decisions on the current token states. 
Following the official implementation of dKV-Cache-Decode~\citep{ma2025dkvcache}, we evaluate dKV-Cache-Decode with a cache refresh step $s=8$, and block size $B=\texttt{32}$.
To achieve parallel decoding, we commit the two tokens with the highest predicted confidence at each denoising step.
For the Dream model, following the official codebase, we apply an entropy-based remasking strategy and set the cache refresh step to $s=4$.

\paragraph{Elastic-Cache}
Elastic-Cache is a training-free KV caching method for diffusion LLMs that adaptively decides when to refresh cached states and from which layer recomputation should resume~\citep{nguyen-tri2026attention}. 
By using attention-aware drift to invalidate caches, Elastic-Cache reuses stable shallow-layer and off-window \texttt{MASK} caches instead of recomputing all cache states at every denoising step.
Following the official implementation of Elastic-Cache~\citep{nguyen-tri2026attention}, we evaluate Elastic-Cache with a refresh threshold of 0.9, a window length of 32, a confidence threshold of 0.9, and a tracked-token count of 1.

\paragraph{WINO}
WINO achieves revocable parallel decoding via a wide-in narrow-out draft-and-verify mechanism~\citep{hong2026widein}.
During the draft stage, multiple candidate tokens are unmasked in parallel. 
The verification stage evaluates each drafted token under full bidirectional context, identifying unreliable predictions. 
Tokens that fail verification are re-masked and refined in subsequent denoising steps, while reliable tokens remain committed. 
This mechanism allows for aggressive parallel decoding without permanently committing to erroneous predictions.
Following the official implementation of WINO~\citep{hong2026widein}, we evaluate WINO with a forward confidence threshold of 0.6, a backward verification threshold of 0.9, and an acceptance window of 30 tokens.
Additionally, consistent with the original implementation, we set the block size to half of the generation length.

\subsection{Implementation details }
\label{implementation}
% \paragraph{Evaluation protocol}
% In the main experiments, we evaluate the performance of current decoding methods on math and code benchmarks.
% To eliminate the effects of hardware and system configuration on evaluation results, our experiments are conducted on a single machine with a single GPU and a batch size of 1.
% All experiments are conducted on L40 GPUs unless otherwise stated. 
% Experiments across hardware platforms are conducted on L40, Pro6000, A100, and RTX 4090 GPUs.
% The prompt templates used in our evaluation are sourced from lm-evaluation-harness and a set of manually crafted prompts, as presented in the Appendix~\ref{total_prompts}.
% For mathematical datasets, we employ Math-Verify to assess the accuracy of generated answers.
% For code datasets, following prior work~\citep{lu2026semanticaware}, we employ the Evaluate module from Hugging Face Evaluate to assess the correctness of generated solutions.
% Unless otherwise specified, for the main experiments, we set the generation length to 128 and adopt the semi-autoregressive decoding strategy with a block size of 32.

\paragraph{Evaluation protocol}
In the main experiments, we evaluate the performance of current decoding methods on math and code benchmarks.
To eliminate the effects of hardware and system configuration on evaluation results, our experiments are conducted on a single machine with a single GPU and a batch size of 1.
All experiments are conducted on L40 GPUs unless otherwise stated.
Experiments across hardware platforms are conducted on L40, Pro6000, A100, and RTX 4090 GPUs.
The prompt templates used in our evaluation are sourced from lm-evaluation-harness and a set of manually crafted prompts (Appendix~\ref{total_prompts}).
For mathematical datasets, we employ Math-Verify to assess the accuracy of generated answers.
For code datasets, following prior work~\citep{lu2026semanticaware}, we employ the Evaluate module from Hugging Face Evaluate to assess the correctness of generated solutions.
Unless otherwise specified, for the main experiments, we set the generation length to 128 and adopt the semi-autoregressive decoding strategy with a block size of 32.

\paragraph{Ensuring reproducibility across multiple runs}
Our work aims to investigate the evaluation inconsistency in dLLMs. 
We first adopt rigorous experimental controls to eliminate the effect of computational randomness, and ensure that evaluation results are reproducible under identical configurations.
Specifically, we adopt the following experimental settings: \textbf{(1) Deterministic computation.} We control sources of randomness in Python, NumPy, and PyTorch, and enforce deterministic computation in cuDNN. \textbf{(2) Consistent hardware platform.} All methods are evaluated on the same GPU to avoid variation caused by hardware differences. \textbf{(3) Single-GPU evaluation with a batch size of 1.} Each evaluation is conducted on a single GPU with a batch size of 1 to avoid nondeterminism introduced by parallel execution. \textbf{(4) Deterministic decoding.} All decoding methods are evaluated using deterministic decoding without stochastic sampling.
% Under these controlled settings, we find that three repeated evaluations under the same prompt template yield identical accuracy, indicating zero variance across runs.
Under these controlled settings, repeated evaluations under the same prompt template produce identical results, confirming that the performance variations analyzed in our experiments are not caused by run-to-run computational randomness.

% \subsection{The definition of Kendall's tau}
% \label{Kendall}
% Kendall's tau quantifies the agreement between two rankings. Given two rankings
% $r$ and $r'$ over the same set of $n$ items, let $N_c$ and $N_d$ be the numbers
% of concordant and discordant item pairs, respectively. We define Kendall's tau as:
% \begin{equation}
% \tau(r, r') = \frac{N_c - N_d}{\binom{n}{2}}.
% \end{equation}
% Higher values indicate stronger agreement between the two rankings, whereas lower values indicate greater disagreement. 
% In our work, we use Kendall’s tau to quantify ranking consistency across prompt templates, which serves as a metric for measuring evaluation inconsistency in comparison of decoding methods.

\subsection{The Definition of Kendall's Tau}
\label{Kendall}
Kendall's tau quantifies the agreement between two rankings.
Given two rankings $r$ and $r'$ over the same set of $n$ items,
let $N_c$ and $N_d$ denote the numbers of concordant and
discordant pairs, respectively. We define Kendall's $\tau_b$ as
\begin{equation}
\tau_b(r,r')
=
\frac{N_c-N_d}
{
\sqrt{
\left(N_0-T_r\right)
\left(N_0-T_{r'}\right)
}
},
\end{equation}
where $N_0=\binom{n}{2}$ denotes the total number of item pairs,
and $T_r$ and $T_{r'}$ denote the numbers of tied pairs in
rankings $r$ and $r'$, respectively. 
Higher values indicate stronger agreement between the two rankings, whereas lower
values indicate greater disagreement.

\subsection{The definition of token-level edit similarity}
\label{edit_similarity}
To quantify response variation across prompt templates, we compute the pairwise token-level edit similarity between generated responses. 
Given two responses, $r$ and $r'$, generated for the same query with different prompt templates, let $D_{\mathrm{Lev}}(r, r')$ denote their token-level Levenshtein distance, which is the minimum number of insertions, deletions, and substitutions of a single token required to transform $r$ into $r'$.
Formally, the token-level edit similarity is defined as
\begin{equation}
S(r,r')
=
1-
\frac{D_{\mathrm{Lev}}(r,r')}
{\max(|r|,|r'|)},
\end{equation}
where $|r|$ and $|r'|$ denote the number of tokens in the two responses, respectively. 
The similarity score $S(r,r') \in [0,1]$, where $S(r,r')=1$ indicates that $r$ and $r'$ are identical sequences, while smaller values indicate greater token-level differences.

\section{Experimental Results}
% \subsection{Decoding methods exhibit pervasive evaluation inconsistency}
% \label{Kendall}
% Figure~\ref{Kendall's_llada8b} and Figure~\ref{Kendall's_llada1.5} show the pairwise Kendall's tau rank correlations among method rankings obtained under different prompt templates on GSM8K for LLaDA-8B-Instruct and LLaDA-1.5, respectively. Higher values indicate more consistent relative rankings of decoding methods, whereas lower values reflect greater disagreement in method rankings across prompts. These results suggest that relying on a single prompt template may produce unreliable performance metrics and yield evaluation illusions for dLLM decoding methods.
% \begin{figure*}[t!]
%     \centering
%     \begin{subfigure}[t]{0.45\textwidth}
%         \centering
%         \includegraphics[width=\linewidth]{fig/Kendall's tau1.pdf}
%         \caption{LLaDA-8B-Instruct}
%         \label{Kendall's_llada8b}
%     \end{subfigure}
%     \hspace{0.04cm} % 调整左右间距
%     \begin{subfigure}[t]{0.45\textwidth}
%         \centering
%         \includegraphics[width=\linewidth]{fig/Kendall's tau2.pdf}
%         \caption{LLaDA-1.5}
%         \label{Kendall's_llada1.5}
%     \end{subfigure}
%     \caption{Pairwise Kendall's tau rank correlation matrices across prompt templates on GSM8K for (a) LLaDA-8B-Instruct and (b) LLaDA-1.5. Each cell quantifies agreement in the rankings of decoding methods across two prompt templates, with higher values indicating more consistent rankings.}
% \end{figure*}

% \input{table/code_dream}

\subsection{Ranking stability of decoding methods across different prompts}
\label{ranking}
\begin{figure*}[t!]
    \centering
    \begin{subfigure}[t]{0.45\textwidth}
        \centering
        \includegraphics[width=\linewidth]{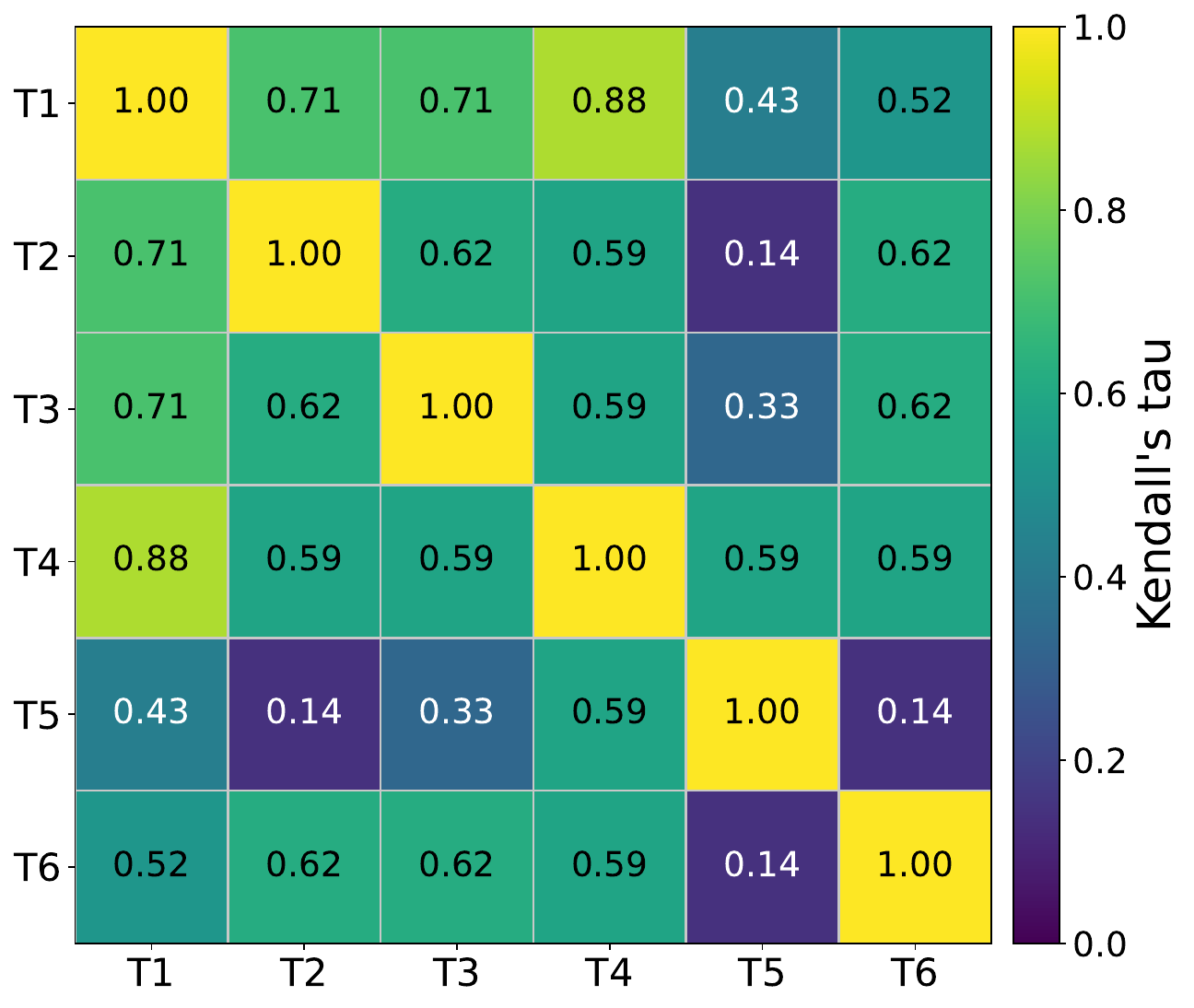}
        \caption{LLaDA-8B-Instruct}
        \label{Kendall's_llada8b}
    \end{subfigure}
    \hspace{0.04cm} % 调整左右间距
    \begin{subfigure}[t]{0.45\textwidth}
        \centering
        \includegraphics[width=\linewidth]{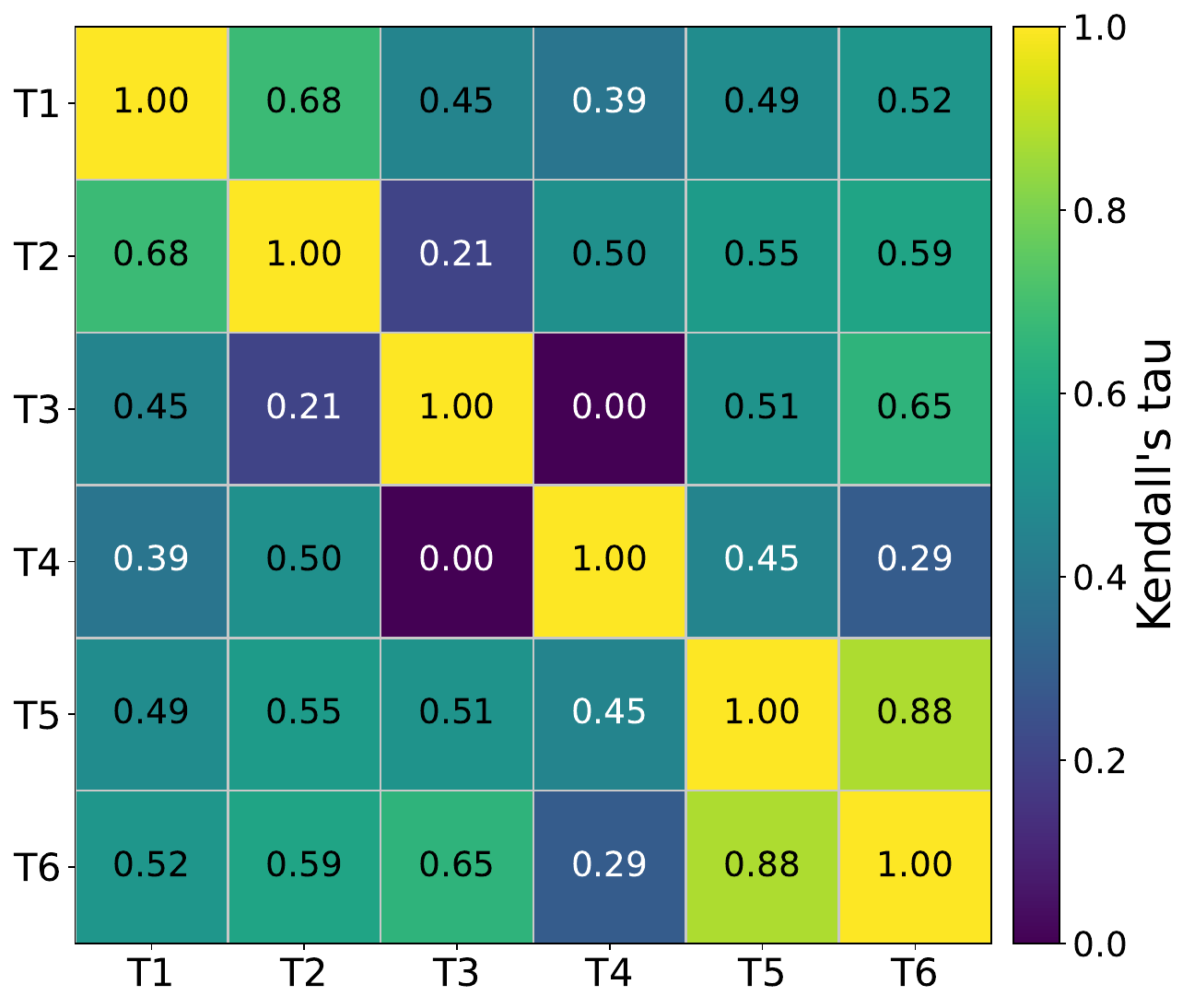}
        \caption{LLaDA-1.5}
        \label{Kendall's_llada1.5}
    \end{subfigure}
    \caption{Pairwise Kendall's tau correlation matrices across prompt templates on GSM8K for (a) LLaDA-8B-Instruct and (b) LLaDA-1.5. Each cell quantifies the ranking agreement of decoding methods between two prompt templates, with higher values indicating more consistent rankings.}
\end{figure*}

Figure~\ref{Kendall's_llada8b} and Figure~\ref{Kendall's_llada1.5} show the pairwise Kendall's tau correlations (See Definition in Appendix~\ref{Kendall}) among method rankings obtained under different prompt templates for LLaDA-8B-Instruct and LLaDA-1.5, respectively.
The average pairwise Kendall's tau across prompt templates is only 0.539 for LLaDA-8B-Instruct and 0.477 for LLaDA-1.5, indicating pervasive evaluation inconsistency across dLLM decoding methods.

\subsection{Statistical confidence analysis with bootstrap resampling}
\label{tab:bootstrap}
To quantify statistical uncertainty, we conduct example level bootstrap sampling with 1,000 iterations on GSM8K.
For each run, we resample examples with replacement using identical indices across all decoding methods and prompt templates to ensure fair comparisons.
We then recompute the average pairwise Kendall's $\tau$ across the prompt templates, reporting 95\% percentile bootstrap confidence intervals.
The average pairwise Kendall's $\tau$ is 0.539 with a 95\% confidence interval of $[0.224, 0.619]$ for LLaDA-8B-Instruct, and 0.477 with a 95\% confidence interval of $[0.206, 0.537]$ for LLaDA-1.5.
The results show that even the upper bounds of the 95\% confidence intervals do not exceed 0.62, indicating that evaluation inconsistency remains pronounced under sample variability.
These findings confirm that the observed evaluation inconsistency across prompts is widespread rather than an artifact of sample resampling.

\subsection{Experimental results on code benchmarks}
\label{code_benchmark}
Table~\ref{tab:math_llada1.5_results} reports the accuracy of vanilla and current decoding methods on GSM8K and MATH500 under LLaDA-1.5.
Table~\ref{tab:code_llada_results} and Table~\ref{tab:code_llada1.5_results} report the accuracy of various decoding methods on code datasets across various prompt templates under LLaDA-8B-Instruct and LLaDA-1.5, respectively.
The results show that current decoding methods achieve occasional marginal improvements over the vanilla baseline on certain prompt templates. However, degradation is widespread across the majority of prompts.
These findings show that current decoding methods do not consistently preserve accuracy across prompt templates while accelerating generation.
\begin{table}[t]
\centering
\caption{The accuracy of vanilla and current decoding methods on GSM8K and MATH500 under LLaDA-1.5, across various prompt templates. \textit{Vanilla} baseline denotes the single-token decoding method. Each numerical subscript indicates the \textcolor{mydarkgreen}{improvement} or \textcolor{mydarkred}{degradation} compared to the vanilla decoding method.}
\label{tab:math_llada1.5_results}
\renewcommand\arraystretch{1.5}
\resizebox{\textwidth}{!}{
\setlength{\tabcolsep}{0.8mm}{
\begin{tabular}{lllllllll|l}
\toprule
\multirow{2}{*}{Method} & \multicolumn{6}{c}{GSM8K} & \multicolumn{2}{c|}{MATH500} & \multicolumn{1}{c}{\multirow{2}{*}{Avg.}} \\
\cmidrule(lr){2-7} \cmidrule(lr){8-9}
& Template 1 & Template 2 & Template 3 & Template 4 & Template 5 & Template 6 & Template 7 & Template 8 & \\
\midrule
\rowcolor{gray!10}
Vanilla
& \underline{78.54} & \textbf{77.86} & \underline{73.92} & \underline{78.01} & \textbf{77.79} & \underline{78.32} & \textbf{38.20} & 33.40 & 67.01 \\
\midrule
\textit{Parallel decoding methods} \\
Fast-dLLM
& \score{77.33}{\down{-1.21}}
& \score{76.80}{\down{-1.06}}
& \score{72.55}{\down{-1.37}}
& \score{\textbf{78.62}}{\up{+0.61}}
& \score{\underline{77.26}}{\down{-0.53}}
& \score{73.77}{\down{-4.55}}
& \score{34.60}{\down{-3.60}}
& \score{\underline{33.60}}{\up{+0.20}}
& \score{65.57}{\down{-1.44}} \\

AdaBlock-dLLM
& \score{77.63}{\down{-0.91}}
& \score{\underline{77.18}}{\down{-0.68}}
& \score{\textbf{74.15}}{\up{+0.23}}
& \score{76.04}{\down{-1.97}}
& \score{\textbf{77.79}}{\up{+0.00}}
& \score{\textbf{78.62}}{\up{+0.30}}
& \score{\underline{37.60}}{\down{-0.60}}
& \score{33.00}{\down{-0.40}}
& \score{66.50}{\down{-0.51}} \\

AdaBlock-Fast
& \score{78.09}{\down{-0.45}}
& \score{\textbf{77.86}}{\up{+0.00}}
& \score{71.80}{\down{-2.12}}
& \score{\underline{78.01}}{\up{+0.00}}
& \score{76.50}{\down{-1.29}}
& \score{73.54}{\down{-4.78}}
& \score{34.20}{\down{-4.00}}
& \score{32.20}{\down{-1.20}}
& \score{65.28}{\down{-1.73}} \\

dKV-Cache-Decode
& \score{75.44}{\down{-3.10}}
& \score{73.84}{\down{-4.02}}
& \score{70.28}{\down{-3.64}}
& \score{72.78}{\down{-5.23}}
& \score{72.25}{\down{-5.54}}
& \score{71.04}{\down{-7.28}}
& \score{29.60}{\down{-8.60}}
& \score{31.80}{\down{-1.60}}
& \score{62.13}{\down{-4.88}} \\

Elastic-Cache
& \score{\textbf{79.23}}{\up{+0.69}}
& \score{76.88}{\down{-0.98}}
& \score{\textbf{74.15}}{\up{+0.23}}
& \score{77.94}{\down{-0.07}}
& \score{76.72}{\down{-1.07}}
& \score{76.04}{\down{-2.28}}
& \score{34.80}{\down{-3.40}}
& \score{\textbf{34.60}}{\up{+1.20}}
& \score{66.30}{\down{-0.71}} \\

WINO
& \score{76.65}{\down{-1.89}}
& \score{76.72}{\down{-1.14}}
& \score{\underline{73.92}}{\up{+0.00}}
& \score{73.09}{\down{-4.92}}
& \score{72.78}{\down{-5.01}}
& \score{73.46}{\down{-4.86}}
& \score{33.80}{\down{-4.40}}
& \score{32.00}{\down{-1.40}}
& \score{64.05}{\down{-2.96}} \\
\bottomrule
\end{tabular}
}}
\end{table}

\begin{table}[t]
\centering
\caption{The accuracy of vanilla and current decoding methods on HumanEval and MBPP under LLaDA-8B-Instruct, across various prompt templates. \textit{Vanilla} baseline denotes the single-token decoding method. Each numerical subscript indicates the \textcolor{mydarkgreen}{improvement} or \textcolor{mydarkred}{degradation} compared to the vanilla decoding method.}
\label{tab:code_llada_results}
\renewcommand\arraystretch{1.5}
\resizebox{\textwidth}{!}{
\setlength{\tabcolsep}{0.8mm}{
\begin{tabular}{lllllllll}
\toprule
\multirow{2}{*}{Method} & \multicolumn{3}{c}{HumanEval} & \multicolumn{3}{c}{MBPP} & \multicolumn{1}{c}{\multirow{2}{*}{Avg.}} \\
\cmidrule(lr){2-4} \cmidrule(lr){5-7}
& Template 1 & Template 2 & Template 3 & Template 4 & Template 5 & Template 6 & \\
\midrule

\rowcolor{gray!10}
Vanilla
& \textbf{35.40} & 29.90 & \textbf{33.50}
& 36.60 & \textbf{40.20} & 37.60
& \underline{35.53} \\
\midrule
\textit{Parallel decoding methods} \\

Fast-dLLM
& \score{\textbf{35.40}}{\up{+0.00}}
& \score{\textbf{32.90}}{\up{+3.00}}
& \score{31.70}{\down{-1.80}}
& \score{37.40}{\up{+0.80}}
& \score{36.40}{\down{-3.80}}
& \score{36.80}{\down{-0.80}}
& \score{35.10}{\down{-0.43}} \\

AdaBlock-dLLM
& \score{34.10}{\down{-1.30}}
& \score{31.70}{\up{+1.80}}
& \score{32.90}{\down{-0.60}}
& \score{37.20}{\up{+0.60}}
& \score{\underline{40.00}}{\down{-0.20}}
& \score{\textbf{38.20}}{\up{+0.60}}
& \score{\textbf{35.68}}{\up{+0.15}} \\

AdaBlock-Fast
& \score{32.90}{\down{-2.50}}
& \score{29.30}{\down{-0.60}}
& \score{\textbf{33.50}}{\up{+0.00}}
& \score{\underline{37.60}}{\up{+1.00}}
& \score{38.80}{\down{-1.40}}
& \score{\underline{38.00}}{\up{+0.40}}
& \score{35.02}{\down{-0.52}} \\

dKV-Cache-Decode
& \score{23.80}{\down{-11.60}}
& \score{20.70}{\down{-9.20}}
& \score{26.20}{\down{-7.30}}
& \score{24.80}{\down{-11.80}}
& \score{29.00}{\down{-11.20}}
& \score{27.00}{\down{-10.60}}
& \score{25.25}{\down{-10.28}} \\

Elastic-Cache
& \score{32.30}{\down{-3.10}}
& \score{\underline{32.30}}{\up{+2.40}}
& \score{29.90}{\down{-3.60}}
& \score{\textbf{39.40}}{\up{+2.80}}
& \score{37.20}{\down{-3.00}}
& \score{36.20}{\down{-1.40}}
& \score{34.55}{\down{-0.98}} \\

WINO
& \score{25.00}{\down{-10.40}}
& \score{25.60}{\down{-4.30}}
& \score{28.00}{\down{-5.50}}
& \score{36.40}{\down{-0.20}}
& \score{35.40}{\down{-4.80}}
& \score{36.00}{\down{-1.60}}
& \score{31.07}{\down{-4.47}} \\

\bottomrule
\end{tabular}
}}
\end{table}
\begin{table}[t]
\centering
\caption{The accuracy of vanilla and current decoding methods on HumanEval and MBPP under LLaDA-1.5, across various prompt templates. \textit{Vanilla} baseline denotes the single-token decoding method. Each numerical subscript indicates the \textcolor{mydarkgreen}{improvement} or \textcolor{mydarkred}{degradation} compared to the vanilla decoding method.}
\label{tab:code_llada1.5_results}
\renewcommand\arraystretch{1.5}
\resizebox{\textwidth}{!}{
\setlength{\tabcolsep}{0.8mm}{
\begin{tabular}{lllllllll}
\toprule
\multirow{2}{*}{Method} & \multicolumn{3}{c}{HumanEval} & \multicolumn{3}{c}{MBPP} & \multicolumn{1}{c}{\multirow{2}{*}{Avg.}} \\
\cmidrule(lr){2-4} \cmidrule(lr){5-7}
& Template 1 & Template 2 & Template 3 & Template 4 & Template 5 & Template 6 & \\
\midrule

\rowcolor{gray!10}
Vanilla
& \textbf{42.10} & \underline{38.40} & \textbf{43.30}
& \textbf{39.60} & \textbf{41.40} & \textbf{40.80}
& \textbf{40.93} \\
\midrule
\textit{Parallel decoding methods} \\

Fast-dLLM
& \score{36.00}{\down{-6.10}}
& \score{32.90}{\down{-5.50}}
& \score{38.40}{\down{-4.90}}
& \score{38.20}{\down{-1.40}}
& \score{40.00}{\down{-1.40}}
& \score{38.60}{\down{-2.20}}
& \score{37.35}{\down{-3.58}} \\

AdaBlock-dLLM
& \score{\underline{39.00}}{\down{-3.10}}
& \score{\textbf{39.60}}{\up{+1.20}}
& \score{\underline{42.70}}{\down{-0.60}}
& \score{\underline{38.60}}{\down{-1.00}}
& \score{39.80}{\down{-1.60}}
& \score{\underline{40.60}}{\down{-0.20}}
& \score{\underline{40.05}}{\down{-0.88}} \\

AdaBlock-Fast
& \score{36.00}{\down{-6.10}}
& \score{33.50}{\down{-4.90}}
& \score{38.40}{\down{-4.90}}
& \score{37.20}{\down{-2.40}}
& \score{\underline{40.60}}{\down{-0.80}}
& \score{39.80}{\down{-1.00}}
& \score{37.58}{\down{-3.35}} \\

dKV-Cache-Decode
& \score{28.70}{\down{-13.40}}
& \score{28.00}{\down{-10.40}}
& \score{32.90}{\down{-10.40}}
& \score{27.80}{\down{-11.80}}
& \score{31.80}{\down{-9.60}}
& \score{33.00}{\down{-7.80}}
& \score{30.37}{\down{-10.57}} \\

Elastic-Cache
& \score{37.20}{\down{-4.90}}
& \score{36.60}{\down{-1.80}}
& \score{38.40}{\down{-4.90}}
& \score{\underline{38.60}}{\down{-1.00}}
& \score{39.60}{\down{-1.80}}
& \score{39.80}{\down{-1.00}}
& \score{38.37}{\down{-2.57}} \\

WINO
& \score{34.80}{\down{-7.30}}
& \score{33.50}{\down{-4.90}}
& \score{37.80}{\down{-5.50}}
& \score{33.20}{\down{-6.40}}
& \score{36.00}{\down{-5.40}}
& \score{39.20}{\down{-1.60}}
& \score{35.75}{\down{-5.18}} \\

\bottomrule
\end{tabular}
}}
\end{table}

\subsection{Experimental results on Dream}
\label{dream_math}
We further evaluate Dream-v0-Instruct-7B to investigate whether the observed evaluation inconsistency extends across model architectures.
Since the official WINO~\citep{hong2026widein} codebase does not include an implementation for Dream, we do not report WINO results for this model.
In practice, we adopt the best-performing hyperparameter settings reported in the corresponding paper or provided in the official codebase. 
Table~\ref{tab:math_dream_results} reports the accuracy of various decoding methods on math datasets across various prompt templates under Dream-v0-Instruct-7B.
We find that the rankings of decoding methods can vary across prompt templates. 
Moreover, the results show that although some decoding methods achieve higher accuracy than the vanilla baseline under certain prompt templates, they underperform the baseline across the majority of templates. 
These results suggest that current parallel decoding methods can accelerate generation but can come at a cost to accuracy on downstream tasks.
\begin{table}[t]
\centering
\caption{The accuracy of vanilla and current decoding methods on GSM8K and MATH500 under Dream-v0-Instruct-7B, across various prompt templates. \textit{Vanilla} baseline denotes the single-token decoding method. Each numerical subscript indicates the \textcolor{mydarkgreen}{improvement} or \textcolor{mydarkred}{degradation} compared to the vanilla decoding method.}
\label{tab:math_dream_results}
\renewcommand\arraystretch{1.5}
\resizebox{\textwidth}{!}{
\setlength{\tabcolsep}{0.8mm}{
\begin{tabular}{lllllllll|l}
\toprule
\multirow{2}{*}{Method}
& \multicolumn{6}{c}{GSM8K}
& \multicolumn{2}{c|}{MATH500}
& \multicolumn{1}{c}{\multirow{2}{*}{Avg.}} \\
\cmidrule(lr){2-7}
\cmidrule(lr){8-9}
& Template 1
& Template 2
& Template 3
& Template 4
& Template 5
& Template 6
& Template 7
& Template 8
& \\
\midrule

\rowcolor{gray!10}
Vanilla
& \textbf{76.27}
& \textbf{78.01}
& \textbf{76.19}
& \textbf{72.71}
& \textbf{73.39}
& 65.28
& 39.00
& \textbf{42.60}
& 65.43 \\

\midrule
\textit{Parallel decoding methods} \\

Fast-dLLM
& \score{75.06}{\down{-1.21}}
& \score{76.72}{\down{-1.29}}
& \score{\underline{75.13}}{\down{-1.06}}
& \score{\underline{72.10}}{\down{-0.61}}
& \score{70.96}{\down{-2.43}}
& \score{\textbf{69.83}}{\up{+4.55}}
& \score{\textbf{41.20}}{\up{+2.20}}
& \score{40.00}{\down{-2.60}}
& \score{65.13}{\down{-0.31}} \\

AdaBlock-dLLM
& \score{\underline{75.89}}{\down{-0.38}}
& \score{\underline{77.33}}{\down{-0.68}}
& \score{\textbf{76.19}}{\up{0.00}}
& \score{71.72}{\down{-0.99}}
& \score{\underline{71.80}}{\down{-1.59}}
& \score{64.90}{\down{-0.38}}
& \score{39.00}{\up{0.00}}
& \score{\underline{42.40}}{\down{-0.20}}
& \score{64.90}{\down{-0.53}} \\

AdaBlock-Fast
& \score{74.68}{\down{-1.59}}
& \score{76.72}{\down{-1.29}}
& \score{74.98}{\down{-1.21}}
& \score{70.51}{\down{-2.20}}
& \score{71.11}{\down{-2.28}}
& \score{\underline{68.23}}{\up{+2.95}}
& \score{\underline{40.60}}{\up{+1.60}}
& \score{39.60}{\down{-3.00}}
& \score{64.55}{\down{-0.88}} \\

dKV-Cache-Decode
& \score{67.40}{\down{-8.87}}
& \score{69.75}{\down{-8.26}}
& \score{66.49}{\down{-9.70}}
& \score{64.06}{\down{-8.65}}
& \score{63.00}{\down{-10.39}}
& \score{63.08}{\down{-2.20}}
& \score{32.00}{\down{-7.00}}
& \score{34.60}{\down{-8.00}}
& \score{57.55}{\down{-7.88}} \\

dKV-Cache-Decode (Half)
& \score{73.39}{\down{-2.88}}
& \score{76.42}{\down{-1.59}}
& \score{73.84}{\down{-2.35}}
& \score{68.39}{\down{-4.32}}
& \score{69.29}{\down{-4.10}}
& \score{66.87}{\up{+1.59}}
& \score{38.80}{\down{-0.20}}
& \score{40.60}{\down{-2.00}}
& \score{63.45}{\down{-1.98}} \\

Elastic-Cache
& \score{70.51}{\down{-5.76}}
& \score{70.43}{\down{-7.58}}
& \score{69.67}{\down{-6.52}}
& \score{65.88}{\down{-6.83}}
& \score{65.58}{\down{-7.81}}
& \score{63.38}{\down{-1.90}}
& \score{35.60}{\down{-3.40}}
& \score{36.80}{\down{-5.80}}
& \score{59.73}{\down{-5.70}} \\

\bottomrule
\end{tabular}
}}
\end{table}

% \subsection{Comparison of dLLMs and AR Models.}
% To compare the prompt sensitivity of autoregressive LLMs and dLLMs, we evaluate LLaDA-8B-Instruct and Qwen3-8B on GSM8K. Specifically, we use threshold-based decoding ($\gamma=0.8$) for the dLLM and greedy decoding for the AR LLM, evaluating both models with a set of nearly identical prompt templates, as detailed in Appendix~\ref{prompt_subtle}. The table~\ref{tab:dllm_ar} below reports the accuracy of LLaDA-8B and Qwen3-8B on GSM8K across different prompt templates. LLaDA-8B exhibits a higher standard deviation and a wider range than Qwen3-8B, indicating that dLLM is more sensitive to prompt template variations.
% \input{table/ar}

\subsection{Sensitivity of decoding methods to prompt templates}
\label{minior_prompt}
Table~\ref{tab:prompt} reports the accuracy of the decoding methods across various near-identical prompt templates on GSM8K under LLaDA-8B-Instruct. 
In practice, we quantify the sensitivity of decoding methods to prompt variation using the standard deviation and range of accuracy across prompt templates.
The results show that parallel decoding methods exhibit substantially greater sensitivity to minor variations of the prompt template.
\begin{table}[th]
\centering
\caption{Accuracy comparison across prompt templates (A--H) on the GSM8K dataset. 
Qwen3-8B serves as the autoregressive reference, while LLaDA-8B-Instruct is evaluated under various decoding strategies. 
\textit{Vanilla} denotes the standard single-token decoding baseline. 
\textit{Confidence Threshold} decodes all masked tokens whose confidence exceeds $\gamma$ at each step. 
\textit{Top-K} decodes the top $k$ most confident masked tokens per step. 
\textbf{Avg} denotes the mean accuracy across templates. 
\textbf{Std} and \textbf{Range} denote the sample standard deviation and range of accuracy across templates, quantifying sensitivity to prompt variation.}
\label{tab:prompt}

\renewcommand{\arraystretch}{1.3}
\resizebox{\textwidth}{!}{
\setlength{\tabcolsep}{1mm}
\begin{tabular}{llcccccccc|ccc}
\toprule
\textbf{Decoding Method} & \textbf{Model}
& \textbf{A} & \textbf{B} & \textbf{C} & \textbf{D}
& \textbf{E} & \textbf{F} & \textbf{G} & \textbf{H}
& \textbf{Avg} & \textbf{Std} & \textbf{Range} \\
\midrule

\textbackslash
& AR
& 91.74 & 91.96 & 91.36 & 91.74
& 91.81 & 91.89 & 91.74 & 91.05
& 91.66 & 0.304 & 0.910 \\

\midrule

Vanilla
& dLLM
& 77.26 & 77.63 & 77.10 & 78.17
& 77.26 & 77.86 & 77.94 & 77.86
& 77.64 & 0.387 & 1.070 \\

\midrule
\textit{Threshold-based decoding methods} \\
Confidence Threshold ($\gamma=0.9$)
& \multirow{2}{*}{dLLM}
& 77.48 & 77.94 & 77.03 & 78.32
& 76.95 & 77.86 & 77.63 & 77.71
& 77.62 & 0.459 & 1.370 \\

Confidence Threshold ($\gamma=0.8$)
&
& 76.19 & 76.19 & 76.42 & 77.63
& 76.27 & 77.48 & 77.33 & 76.65
& 76.77 & 0.611 & 1.440 \\

\midrule
\textit{Top-K decoding methods} \\
Top-K ($k=2$)
& \multirow{2}{*}{dLLM}
& 75.36 & 73.69 & 74.45 & 74.83
& 75.06 & 74.68 & 74.00 & 72.18
& 74.28 & 1.007 & 3.180 \\

Top-K ($k=4$)
&
& 65.58 & 66.64 & 66.57 & 68.23
& 66.19 & 65.05 & 65.43 & 64.44
& 66.02 & \textbf{1.170} & \textbf{3.790} \\

\bottomrule
\end{tabular}
}
\end{table}

\subsection{Effect of prompt variations on generation trajectories}
\label{divergence}
\begin{figure}[t!]
    \centering
    \includegraphics[
        width=0.9\linewidth,
        height=0.3\textheight
    ]{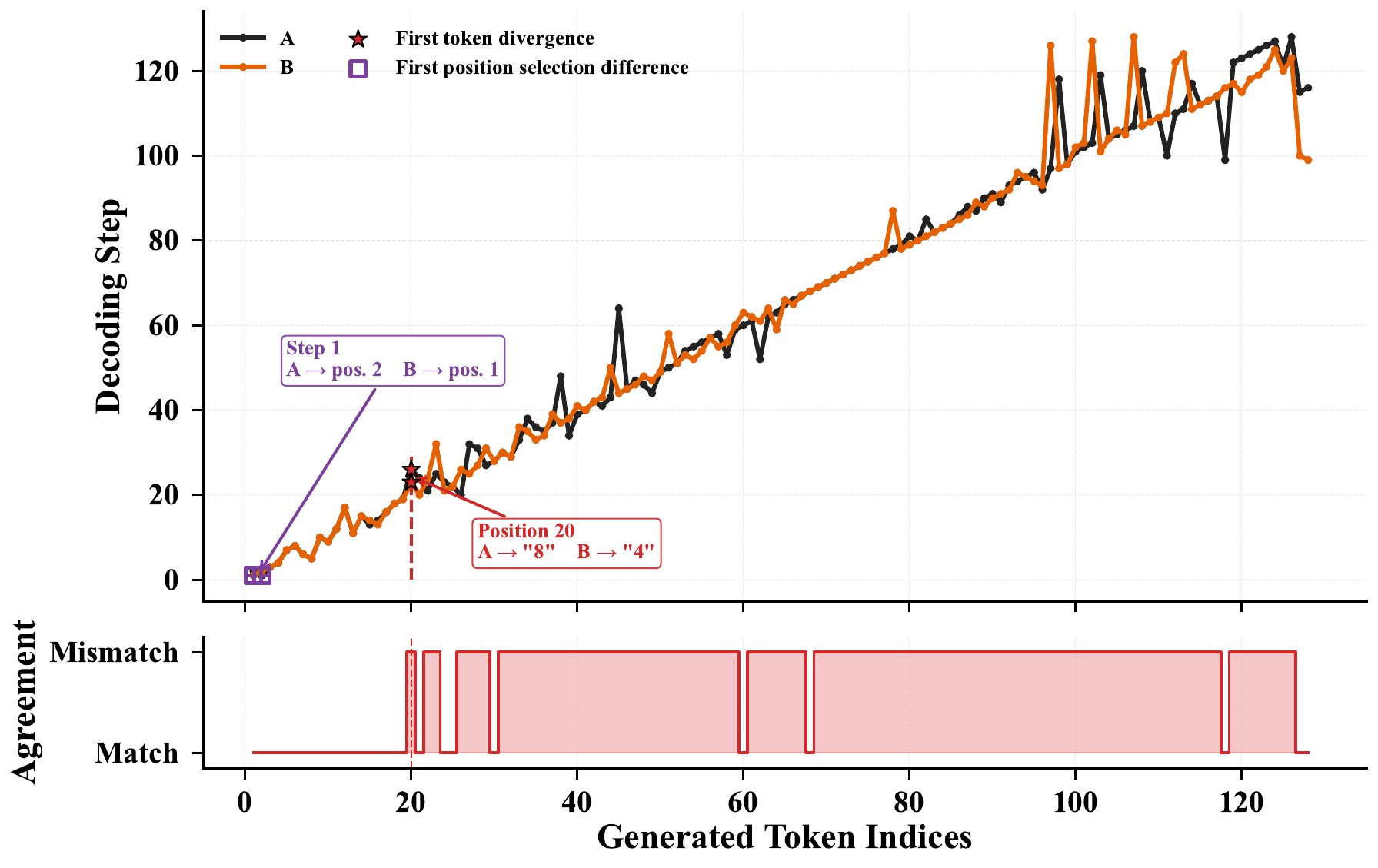}
    \caption{Illustration of prompt-induced decoding divergence.
    (Top) Comparison of decoding dynamics between two prompt templates:
    A difference in position selection already emerges at the first decoding step.
    (Bottom) Position-wise token agreement between the two generated sequences: the first token mismatch occurs at position 20.}
    \label{fig:generation}
    % \vspace{-3mm}
\end{figure}

In Figure~\ref{fig:generation}, we illustrate how subtle prompt variations induce decoding dynamics in dLLMs.
We find that a difference in position selection already emerges at the first decoding step, where prompt A chooses position 2, whereas prompt B chooses position 1.
The first token mismatch occurs later at position 20, after which token-level disagreements become increasingly frequent at subsequent decoding positions.
dLLMs often determine both what tokens to generate and where to unmask at each decoding step.
This flexibility makes the decoding trajectory jointly dependent on token prediction and position selection, thereby increasing sensitivity to subtle prompt variations.
As shown in Example~\ref{box:discrepancies}, we present a detailed case study illustrating how minor prompt changes can alter sequence generation. 
We find that even a subtle variation in the prompt can induce divergent reasoning trajectories and yield different final answers.

\subsection{Empirical analysis of the speed--accuracy trade-off}
\label{trade-off}
Current dLLM decoding methods substantially improve generation efficiency, which is critical to the practical deployment of dLLMs. Tables~\ref{tab:math_llada_steps} and~\ref{tab:math_llada1.5_steps} report the number of function evaluations (NFE) for decoding methods whose accuracy is presented in Tables~\ref{tab:math_llada_results} and~\ref{tab:math_llada1.5_results}, respectively. 
The NFE metric measures the number of model forward passes performed during decoding. 
The experimental results show that existing methods reduce the number of decoding steps and thereby improve generation efficiency.
Additionally, Table~\ref{tab:llada8b_speed_math} and Table~\ref{tab:llada1_5b_speed} show the accuracy and decoding speedup of current decoding methods on LLaDA-8B-Instruct and LLaDA-1.5, across various prompt templates. The speedup is measured relative to the decoding latency of Vanilla under the same evaluation configuration. 
The experimental results show that the performance gains of parallel decoding methods are highly prompt-dependent, and that accelerating generation can degrade downstream task accuracy.

\begin{table}[t]
\centering
\caption{The decoding steps of vanilla and parallel decoding methods on GSM8K and MATH500 under LLaDA-8B-Instruct, across various evaluation prompt templates. \textit{Vanilla} denotes the low-confidence decoding baseline, which decodes a single masked token with the highest confidence at each step.}
\label{tab:math_llada_steps}
\renewcommand\arraystretch{1.5}
\resizebox{\textwidth}{!}{
\setlength{\tabcolsep}{0.8mm}{
\begin{tabular}{lllllllll|l}
\toprule
\multirow{2}{*}{Method} & \multicolumn{6}{c}{GSM8K} & \multicolumn{2}{c|}{MATH500} & \multicolumn{1}{c}{\multirow{2}{*}{Avg.}} \\
\cmidrule(lr){2-7} \cmidrule(lr){8-9}
& Template 1 & Template 2 & Template 3 & Template 4 & Template 5 & Template 6 & Template 7 & Template 8 & \\
\midrule

\rowcolor{gray!10}
Vanilla
& 128 & 128 & 128 & 128 & 128 & 128 & 128 & 128 & 128.00 \\
\midrule
\textit{Parallel decoding methods} \\

Fast-dLLM
& 57
& 60
& 56
& 54
& 54
& 53
& 69
& 66
& 58.63 \\

AdaBlock-dLLM
& 53
& 56
& 52
& 50
& 50
& 48
& 65
& 62
& 54.50 \\

AdaBlock-Fast
& 57
& 61
& 57
& 55
& 55
& 55
& 70
& 67
& 59.63 \\

dKV-Cache-Decode
& 64
& 64
& 64
& 64
& 64
& 64
& 64
& 64
& 64.00 \\

Elastic-Cache
& 54
& 57
& 53
& 50
& 51
& 49
& 68
& 63
& 55.63 \\

WINO
& 31
& 32
& 31
& 27
& 27
& 27
& 44
& 37
& 32.00 \\

\bottomrule
\end{tabular}
}}
\end{table}
\begin{table}[t]
\centering
\caption{The decoding steps of vanilla and parallel decoding methods on GSM8K and MATH500 under LLaDA-1.5, across various prompt templates. \textit{Vanilla} denotes the low-confidence decoding baseline, which decodes a single masked token with the highest confidence at each step.}
\label{tab:math_llada1.5_steps}
\renewcommand\arraystretch{1.5}
\resizebox{\textwidth}{!}{
\setlength{\tabcolsep}{0.8mm}{
\begin{tabular}{lllllllll|l}
\toprule
\multirow{2}{*}{Method} & \multicolumn{6}{c}{GSM8K} & \multicolumn{2}{c|}{MATH500} & \multicolumn{1}{c}{\multirow{2}{*}{Avg.}} \\
\cmidrule(lr){2-7} \cmidrule(lr){8-9}
& Template 1 & Template 2 & Template 3 & Template 4 & Template 5 & Template 6 & Template 7 & Template 8 & \\
\midrule

\rowcolor{gray!10}
Vanilla
& 128 & 128 & 128 & 128 & 128 & 128 & 128 & 128 & 128.00 \\
\midrule
\textit{Parallel decoding methods} \\

Fast-dLLM
& 53
& 58
& 55
& 53
& 53
& 49
& 68
& 65
& 56.75 \\

AdaBlock-dLLM
& 51
& 54
& 52
& 49
& 49
& 46
& 64
& 60
& 53.13 \\

AdaBlock-Fast
& 54
& 59
& 56
& 54
& 54
& 53
& 68
& 66
& 58.00 \\

dKV-Cache-Decode
& 64
& 64
& 64
& 64
& 64
& 64
& 64
& 64
& 64.00 \\

Elastic-Cache
& 54
& 57
& 53
& 50
& 50
& 49
& 66
& 64
& 55.38 \\

WINO
& 30
& 32
& 31
& 28
& 29
& 27
& 43
& 46
& 33.25 \\

\bottomrule
\end{tabular}
}}
\end{table}

% \begin{figure*}[t!]
%     \centering
%     \begin{subfigure}[t]{0.49\textwidth}
%         \centering
%         \includegraphics[width=\linewidth]{fig/pareto_llada.pdf}
%         % \caption{Accuracy across varing thresholds $\gamma$}
%         \caption{}
%         \label{fig:tradeoff_1}
%     \end{subfigure}
%     \hspace{0.1cm} %
%     \begin{subfigure}[t]{0.49\textwidth}
%         \centering
%         \includegraphics[width=\linewidth]{fig/pareto_llada1_5.pdf}
%         % \caption{Accuracy across varying prompts}
%         \caption{}
%         \label{fig:tradeoff_2}
%     \end{subfigure}
%     \caption{As.}
%     % \label{fig:prompt}
% \end{figure*}

\begin{table}[t]
\centering
\caption{The accuracy and decoding speedup of vanilla and current decoding methods on GSM8K and MATH500 under LLaDA-8B-Instruct, across various prompt templates. \textit{Vanilla} denotes the single-token decoding method. For parallel decoding methods, the speedup relative to Vanilla is reported below each accuracy value.}
\label{tab:llada8b_speed_math}
\renewcommand{\arraystretch}{1.5}

\resizebox{\textwidth}{!}{%
\setlength{\tabcolsep}{0.8mm}%
\begin{tabular}{lllllllll|l}
\toprule
\multirow{2}{*}{Method}
& \multicolumn{6}{c}{GSM8K}
& \multicolumn{2}{c|}{MATH500}
& \multicolumn{1}{c}{\multirow{2}{*}{Avg.}} \\
\cmidrule(lr){2-7}
\cmidrule(lr){8-9}
& Template 1
& Template 2
& Template 3
& Template 4
& Template 5
& Template 6
& Template 7
& Template 8
& \\
\midrule

\rowcolor{gray!10}
Vanilla
& \textbf{77.33}
& \underline{74.60}
& 71.72
& \textbf{76.35}
& 74.98
& \underline{76.88}
& \textbf{37.60}
& 34.00
& \textbf{65.43} \\
\midrule

Fast-dLLM
& \shortstack[c]{76.35\\[2pt]{\small (2.46$\times$)}}
& \shortstack[c]{73.16\\[2pt]{\small (2.37$\times$)}}
& \shortstack[c]{70.51\\[2pt]{\small (2.48$\times$)}}
& \shortstack[c]{75.74\\[2pt]{\small (6.85$\times$)}}
& \shortstack[c]{\underline{75.82}\\[2pt]{\small (9.06$\times$)}}
& \shortstack[c]{74.60\\[2pt]{\small (3.55$\times$)}}
& \shortstack[c]{33.80\\[2pt]{\small (2.03$\times$)}}
& \shortstack[c]{\underline{35.20}\\[2pt]{\small (5.83$\times$)}}
& \shortstack[c]{64.40\\[2pt]{\small (4.33$\times$)}} \\
\midrule

AdaBlock-dLLM
& \shortstack[c]{76.88\\[2pt]{\small (2.39$\times$)}}
& \shortstack[c]{\textbf{75.06}\\[2pt]{\small (2.31$\times$)}}
& \shortstack[c]{\underline{71.95}\\[2pt]{\small (2.46$\times$)}}
& \shortstack[c]{75.74\\[2pt]{\small (2.54$\times$)}}
& \shortstack[c]{74.83\\[2pt]{\small (2.56$\times$)}}
& \shortstack[c]{\textbf{76.95}\\[2pt]{\small (2.67$\times$)}}
& \shortstack[c]{\underline{37.00}\\[2pt]{\small (1.78$\times$)}}
& \shortstack[c]{34.40\\[2pt]{\small (1.85$\times$)}}
& \shortstack[c]{\underline{65.35}\\[2pt]{\small (2.32$\times$)}} \\
\midrule

AdaBlock-Fast
& \shortstack[c]{76.65\\[2pt]{\small (2.42$\times$)}}
& \shortstack[c]{73.92\\[2pt]{\small (2.34$\times$)}}
& \shortstack[c]{71.27\\[2pt]{\small (2.43$\times$)}}
& \shortstack[c]{75.66\\[2pt]{\small (6.11$\times$)}}
& \shortstack[c]{75.21\\[2pt]{\small (7.73$\times$)}}
& \shortstack[c]{74.15\\[2pt]{\small (3.37$\times$)}}
& \shortstack[c]{34.20\\[2pt]{\small (2.02$\times$)}}
& \shortstack[c]{\textbf{36.20}\\[2pt]{\small (5.35$\times$)}}
& \shortstack[c]{64.66\\[2pt]{\small (3.97$\times$)}} \\
\midrule

dKV-Cache-Decode
& \shortstack[c]{72.93\\[2pt]{\small (1.98$\times$)}}
& \shortstack[c]{69.45\\[2pt]{\small (2.05$\times$)}}
& \shortstack[c]{66.94\\[2pt]{\small (2.00$\times$)}}
& \shortstack[c]{72.93\\[2pt]{\small (4.27$\times$)}}
& \shortstack[c]{72.63\\[2pt]{\small (4.65$\times$)}}
& \shortstack[c]{73.92\\[2pt]{\small (2.64$\times$)}}
& \shortstack[c]{31.80\\[2pt]{\small (2.04$\times$)}}
& \shortstack[c]{33.00\\[2pt]{\small (4.22$\times$)}}
& \shortstack[c]{61.70\\[2pt]{\small (2.98$\times$)}} \\
\midrule

Elastic-Cache
& \shortstack[c]{\underline{77.10}\\[2pt]{\small \textbf{(2.99$\times$)}}}
& \shortstack[c]{73.84\\[2pt]{\small \textbf{(2.94$\times$)}}}
& \shortstack[c]{\textbf{72.18}\\[2pt]{\small \textbf{(3.05$\times$)}}}
& \shortstack[c]{\underline{75.97}\\[2pt]{\small \textbf{(7.80$\times$)}}}
& \shortstack[c]{\textbf{76.88}\\[2pt]{\small \textbf{(9.64$\times$)}}}
& \shortstack[c]{76.35\\[2pt]{\small \textbf{(4.46$\times$)}}}
& \shortstack[c]{33.60\\[2pt]{\small \textbf{(2.45$\times$)}}}
& \shortstack[c]{31.80\\[2pt]{\small \textbf{(6.59$\times$)}}}
& \shortstack[c]{64.72\\[2pt]{\small \textbf{(4.99$\times$)}}} \\
\midrule

WINO
& \shortstack[c]{74.68\\[2pt]{\small (2.83$\times$)}}
& \shortstack[c]{72.48\\[2pt]{\small (2.77$\times$)}}
& \shortstack[c]{70.74\\[2pt]{\small (2.93$\times$)}}
& \shortstack[c]{74.45\\[2pt]{\small (3.32$\times$)}}
& \shortstack[c]{72.78\\[2pt]{\small (3.07$\times$)}}
& \shortstack[c]{75.21\\[2pt]{\small (3.95$\times$)}}
& \shortstack[c]{33.40\\[2pt]{\small (1.96$\times$)}}
& \shortstack[c]{34.20\\[2pt]{\small (2.25$\times$)}}
& \shortstack[c]{63.49\\[2pt]{\small (2.89$\times$)}} \\

\bottomrule
\end{tabular}%
}
\end{table}
\begin{table}[t]
\centering
\caption{The accuracy and decoding speedup of vanilla and current decoding methods on GSM8K and MATH500 under LLaDA-1.5, across various prompt templates. \textit{Vanilla} denotes the single-token decoding method. For parallel decoding methods, the speedup relative to Vanilla is reported below each accuracy value.}
\label{tab:llada1_5b_speed}
\renewcommand{\arraystretch}{1.5}

\resizebox{\textwidth}{!}{%
\setlength{\tabcolsep}{0.8mm}%
\begin{tabular}{lllllllll|l}
\toprule
\multirow{2}{*}{Method}
& \multicolumn{6}{c}{GSM8K}
& \multicolumn{2}{c|}{MATH500}
& \multicolumn{1}{c}{\multirow{2}{*}{Avg.}} \\
\cmidrule(lr){2-7}
\cmidrule(lr){8-9}
& Template 1
& Template 2
& Template 3
& Template 4
& Template 5
& Template 6
& Template 7
& Template 8
& \\
\midrule

\rowcolor{gray!10}
Vanilla
& \underline{78.54}
& \textbf{77.86}
& \underline{73.92}
& \underline{78.01}
& \textbf{77.79}
& \underline{78.32}
& \textbf{38.20}
& 33.40
& \textbf{67.01} \\
\midrule

Fast-dLLM
& \shortstack[c]{77.33\\[2pt]{\small (2.50$\times$)}}
& \shortstack[c]{76.80\\[2pt]{\small (2.35$\times$)}}
& \shortstack[c]{72.55\\[2pt]{\small (2.46$\times$)}}
& \shortstack[c]{\textbf{78.62}\\[2pt]{\small (6.94$\times$)}}
& \shortstack[c]{\underline{77.26}\\[2pt]{\small (9.20$\times$)}}
& \shortstack[c]{73.77\\[2pt]{\small (3.65$\times$)}}
& \shortstack[c]{34.60\\[2pt]{\small (1.96$\times$)}}
& \shortstack[c]{\underline{33.60}\\[2pt]{\small (5.82$\times$)}}
& \shortstack[c]{65.57\\[2pt]{\small (4.36$\times$)}} \\
\midrule

AdaBlock-dLLM
& \shortstack[c]{77.63\\[2pt]{\small (2.50$\times$)}}
& \shortstack[c]{\underline{77.18}\\[2pt]{\small (2.35$\times$)}}
& \shortstack[c]{\textbf{74.15}\\[2pt]{\small (2.45$\times$)}}
& \shortstack[c]{76.04\\[2pt]{\small (2.56$\times$)}}
& \shortstack[c]{\textbf{77.79}\\[2pt]{\small (2.60$\times$)}}
& \shortstack[c]{\textbf{78.62}\\[2pt]{\small (2.74$\times$)}}
& \shortstack[c]{\underline{37.60}\\[2pt]{\small (1.86$\times$)}}
& \shortstack[c]{33.00\\[2pt]{\small (2.05$\times$)}}
& \shortstack[c]{\underline{66.50}\\[2pt]{\small (2.39$\times$)}} \\
\midrule

AdaBlock-Fast
& \shortstack[c]{78.09\\[2pt]{\small (2.51$\times$)}}
& \shortstack[c]{\textbf{77.86}\\[2pt]{\small (2.35$\times$)}}
& \shortstack[c]{71.80\\[2pt]{\small (2.43$\times$)}}
& \shortstack[c]{\underline{78.01}\\[2pt]{\small (6.35$\times$)}}
& \shortstack[c]{76.50\\[2pt]{\small (8.12$\times$)}}
& \shortstack[c]{73.54\\[2pt]{\small (3.36$\times$)}}
& \shortstack[c]{34.20\\[2pt]{\small (1.97$\times$)}}
& \shortstack[c]{32.20\\[2pt]{\small (5.51$\times$)}}
& \shortstack[c]{65.28\\[2pt]{\small (4.08$\times$)}} \\
\midrule

dKV-Cache-Decode
& \shortstack[c]{75.44\\[2pt]{\small (1.95$\times$)}}
& \shortstack[c]{73.84\\[2pt]{\small (1.99$\times$)}}
& \shortstack[c]{70.28\\[2pt]{\small (1.96$\times$)}}
& \shortstack[c]{72.78\\[2pt]{\small (4.20$\times$)}}
& \shortstack[c]{72.25\\[2pt]{\small (4.63$\times$)}}
& \shortstack[c]{71.04\\[2pt]{\small (2.53$\times$)}}
& \shortstack[c]{29.60\\[2pt]{\small (1.96$\times$)}}
& \shortstack[c]{31.80\\[2pt]{\small (4.24$\times$)}}
& \shortstack[c]{62.13\\[2pt]{\small (2.93$\times$)}} \\
\midrule

Elastic-Cache
& \shortstack[c]{\textbf{79.23}\\[2pt]{\small \textbf{(3.11$\times$)}}}
& \shortstack[c]{76.88\\[2pt]{\small \textbf{(2.92$\times$)}}}
& \shortstack[c]{\textbf{74.15}\\[2pt]{\small \textbf{(3.05$\times$)}}}
& \shortstack[c]{77.94\\[2pt]{\small \textbf{(7.95$\times$)}}}
& \shortstack[c]{76.72\\[2pt]{\small \textbf{(9.84$\times$)}}}
& \shortstack[c]{76.04\\[2pt]{\small \textbf{(4.60$\times$)}}}
& \shortstack[c]{34.80\\[2pt]{\small \textbf{(2.43$\times$)}}}
& \shortstack[c]{\textbf{34.60}\\[2pt]{\small \textbf{(6.68$\times$)}}}
& \shortstack[c]{66.30\\[2pt]{\small \textbf{(5.07$\times$)}}} \\
\midrule

WINO
& \shortstack[c]{76.65\\[2pt]{\small (3.07$\times$)}}
& \shortstack[c]{76.72\\[2pt]{\small (2.88$\times$)}}
& \shortstack[c]{\underline{73.92}\\[2pt]{\small (2.98$\times$)}}
& \shortstack[c]{73.09\\[2pt]{\small (3.20$\times$)}}
& \shortstack[c]{72.78\\[2pt]{\small (2.95$\times$)}}
& \shortstack[c]{73.46\\[2pt]{\small (3.70$\times$)}}
& \shortstack[c]{33.80\\[2pt]{\small (2.11$\times$)}}
& \shortstack[c]{32.00\\[2pt]{\small (2.28$\times$)}}
& \shortstack[c]{64.05\\[2pt]{\small (2.90$\times$)}} \\

\bottomrule
\end{tabular}%
}
\end{table}

\subsection{Experimental results across different prompt templates}
\label{label:acc_step}
Table~\ref{tab:prompt_steps} reports the accuracy and decoding steps of single-token decoding and confidence-threshold decoding strategies on GSM8K under LLaDA-8B-Instruct across a set of semantically equivalent prompt templates. 
The results show that prompt templates can play a dominant role in determining evaluation results. 
An effective template can yield high performance scores even with fewer denoising steps, substantially exceeding the marginal gains from increasing denoising steps.
\begin{table}[t]
\centering
\caption{The accuracy and number of decoding steps of vanilla and current decoding methods on GSM8K datasets under LLaDA-8B-Instruct, across various semantically equivalent prompt 
templates used for evaluation. Vanilla denotes the low-confidence sampling baseline.
}
\label{tab:prompt_steps}
% \small
\renewcommand\arraystretch{1.5}
\resizebox{\textwidth}{!}{
\setlength{\tabcolsep}{2mm}
\begin{tabular}{llcccccccc}
\hline
\textbf{Method} & \textbf{Metric}
& \textbf{A} & \textbf{B} & \textbf{C} & \textbf{D} 
& \textbf{E} & \textbf{F} & \textbf{G} & \textbf{H} \\
\hline
\multirow{2}{*}{Vanilla}
& Acc.  & 77.41 & 76.04 & 72.02 & 71.27 & 72.33 & 73.16 & 68.16 & 78.47 \\
& Step  & 128 & 128 & 128 & 128 & 128 & 128 & 128 & 128 \\
\hline

\multirow{2}{*}{Threshold-based decoding ($\gamma=0.6$)}
& Acc.  & 73.01 & 69.52 & 67.17 & 66.49 & 67.63 & 69.67 & 62.47 & 74.00 \\
& Step  & 24 & 24 & 30 & 25 & 24 & 27 & 28 & 25 \\
\hline

\multirow{2}{*}{Threshold-based decoding ($\gamma=0.7$)}
& Acc.  & 75.97 & 72.63 & 70.05 & 70.36 & 70.13 & 71.95 & 66.57 & 75.97 \\
& Step  & 29 & 29 & 37 & 30 & 29 & 33 & 35 & 31 \\
\hline

\multirow{2}{*}{Threshold-based decoding ($\gamma=0.8$)}
& Acc.  & 76.88 & 75.44 & 71.65 & 70.43 & 71.49 & 72.40 & 67.63 & 77.79 \\
& Step  & 36 & 37 & 45 & 37 & 36 & 41 & 44 & 38 \\
\hline

\multirow{2}{*}{Threshold-based decoding ($\gamma=0.9$)}
& Acc.  & 78.01 & 76.27 & 71.80 & 70.74 & 72.48 & 73.31 & 67.85 & 78.47 \\
& Step  & 45 & 47 & 55 & 46 & 46 & 51 & 57 & 48 \\
\hline
\end{tabular}
}
\end{table}

\paragraph{How does few-shot prompting affect the evaluation of decoding methods?}
\label{few_shots}
Table~\ref{tab:few_shot} reports the accuracy of vanilla and parallel decoding methods across varying few-shot settings on GSM8K under LLaDA-8B-Instruct.
Our experimental results show that the number of few-shot examples can affect the performance of decoding methods during evaluation, suggesting that future evaluations should account for this factor.
\begin{table}[t]
\centering
\caption{The accuracy of vanilla and current decoding methods on GSM8K under LLaDA-8B-Instruct, across different few-shot settings. \textit{Vanilla} denotes the low-confidence sampling baseline. Each numerical subscript indicates the \textcolor{mydarkgreen}{improvement} or \textcolor{mydarkred}{degradation} compared to the vanilla decoding method. Cells highlighted in yellow indicate the best result across few-shot settings for each method.}
\label{tab:few_shot}
\renewcommand\arraystretch{1.2}
\resizebox{\textwidth}{!}{
\setlength{\tabcolsep}{3mm}{
\begin{tabular}{llllll|l}
\toprule
\multirow{2}{*}{Method} & \multicolumn{5}{c|}{GSM8K} & \multicolumn{1}{c}{\multirow{2}{*}{Avg.}} \\
\cmidrule(lr){2-6}
& 0-shot & 1-shot & 2-shot & 3-shot & 4-shot & \\
\midrule

\rowcolor{gray!10}
Vanilla
& 71.72 & 76.04 & \cellcolor{yellow!20}\textbf{76.72} & 76.35 & 75.44 & 75.25 \\

\midrule
\textit{Parallel decoding methods} \\

Fast-dLLM
& \score{70.51}{\down{-1.21}}
& \score{73.77}{\down{-2.27}}
& \score{76.04}{\down{-0.68}}
& \score{75.74}{\down{-0.61}}
& \cellcolor{yellow!20}\score{\underline{76.35}}{\up{+0.91}}
& \score{74.48}{\down{-0.77}} \\

AdaBlock-dLLM
& \score{71.95}{\up{+0.23}}
& \score{74.60}{\down{-1.44}}
& \score{75.82}{\down{-0.91}}
& \score{75.74}{\down{-0.61}}
& \cellcolor{yellow!20}\score{76.12}{\up{+0.68}}
& \score{74.85}{\down{-0.40}} \\

AdaBlock-Fast
& \score{71.27}{\down{-0.45}}
& \score{74.22}{\down{-1.82}}
& \score{75.28}{\down{-1.44}}
& \cellcolor{yellow!20}\score{75.66}{\down{-0.68}}
& \score{75.59}{\up{+0.15}}
& \score{74.40}{\down{-0.85}} \\

dKV-Cache-Decode
& \score{70.28}{\down{-1.44}}
& \score{70.74}{\down{-5.31}}
& \score{72.56}{\down{-4.17}}
& \cellcolor{yellow!20}\score{72.78}{\down{-3.56}}
& \score{72.10}{\down{-3.34}}
& \score{71.69}{\down{-3.56}} \\

Elastic-Cache
& \score{72.18}{\up{+0.45}}
& \score{75.97}{\down{-0.08}}
& \score{76.12}{\down{-0.61}}
& \score{75.97}{\down{-0.38}}
& \cellcolor{yellow!20}\score{\textbf{76.65}}{\up{+1.21}}
& \score{75.38}{\up{+0.14}} \\

WINO
& \score{70.74}{\down{-0.98}}
& \score{73.31}{\down{-2.73}}
& \score{74.75}{\down{-1.97}}
& \score{74.45}{\down{-1.90}}
& \cellcolor{yellow!20}\score{75.13}{\down{-0.31}}
& \score{73.68}{\down{-1.57}} \\

\bottomrule
\end{tabular}
}}
\end{table}

\subsection{Evaluation results with FP32 precision} 
Table~\ref{tab:prompt_gpu} shows the accuracy discrepancies across GPU types under identical decoding methods and prompt templates, evaluated in BF16 precision.
Table~\ref{tab:gpu_float32} presents the accuracy discrepancies between L40 and A100 under the threshold-based decoding method ($\gamma=0.8$) across various prompt templates, evaluated in FP32 precision. The results show that the evaluation results remain consistent between the GPU types at the precision of two decimal places, demonstrating that FP32 precision can substantially reduce evaluation variance across different hardware platforms.
\label{label:float}
\begin{table}[t]
\centering
\caption{Accuracy discrepancies across GPU types under identical decoding methods and prompt templates. 
$|\Delta|$ denotes the absolute accuracy deviation of the other GPU types relative to L40. Avg.\ and Max.\ denote the mean and maximum $|\Delta|$ across all prompt templates, respectively.}
\label{tab:prompt_gpu}
\renewcommand\arraystretch{1.5}
\resizebox{1.0\textwidth}{!}{
\setlength{\tabcolsep}{3mm}{
\begin{tabular}{lcccccccccc}
\toprule
\multirow{2}{*}{GPU} & \multicolumn{8}{c}{Template} & \multirow{2}{*}{Avg.} & \multirow{2}{*}{Max.} \\
\cmidrule(lr){2-9}
 & A & B & C & D & E & F & G & H & & \\
\hline

\hline
\textit{Single-token decoding} \\
\hline
\rowcolor{gray!10}
L40       & 77.26 & 77.63 & 77.10 & 78.17 & 77.26 & 77.86 & 77.94 & 77.86 & -- & -- \\
\hline
Pro6000   & 77.56 & 77.03 & 77.10 & 77.41 & 78.01 & 77.86 & 78.09 & 77.79 & -- & -- \\

% $|\Delta|$ & 0.30 & 0.61 & 0.00 & 0.76 & 0.76 & 0.00 & 0.15 & 0.08 & 0.33 & 0.76 \\
$|\Delta|$ & 0.30 & 0.61 & 0.00 & 0.76 & 0.76 & 0.00 & 0.15 & 0.08 & 0.33 & 0.76 \\
\hline
RTX 4090 & 77.41 & 77.48 & 76.65 & 77.94 & 77.48 & 78.32 & 77.41 & 77.71 \\
$|\Delta|$ & 0.15 & 0.15 & 0.45 & 0.23 & 0.22 & 0.46 & 0.53 & 0.15 & 0.29 & 0.53 \\
\hline
A100 & 77.10 & 77.10 & 76.72 & 77.79 & 76.42 & 77.56 & 78.01 & 78.62 \\
$|\Delta|$ & 0.16 & 0.53 & 0.38 & 0.38 & 0.84 & 0.30 & 0.07 & 0.76 & 0.43 & 0.84 \\
\hline
\textit{Threshold-based parallel decoding ($\gamma$=0.8)} \\
\hline
\rowcolor{gray!10}
L40       & 76.19 & 76.19 & 76.42 & 77.63 & 76.27 & 77.48 & 77.33 & 76.65 & -- & -- \\
\hline
Pro6000   & 76.27 & 76.72 & 76.19 & 77.94 & 76.95 & 77.10 & 76.80 & 77.03 & -- & -- \\
$|\Delta|$ & 0.08 & 0.53 & 0.23 & 0.31 & 0.68 & 0.38 & 0.53 & 0.38 & 0.39 & 0.68 \\
\hline
RTX 4090  & 76.35 & 76.50 & 76.12 & 77.41 & 76.95 & 77.71 & 76.35 & 76.72 & -- & -- \\
$|\Delta|$ & 0.16 & 0.31 & 0.30 & 0.22 & 0.68 & 0.23 & 0.98 & 0.07 & 0.37 & 0.98 \\
\hline
A100 & 76.27 & 76.27 & 75.82 & 77.94 & 76.27 & 77.03 & 76.72 & 77.71 \\
$|\Delta|$ & 0.08 & 0.08 & 0.60 & 0.31 & 0.00 & 0.45 & 0.61 & 1.06 & 0.40 & 1.06 \\

\bottomrule
\end{tabular}
}}
\end{table}

% % Single-token decoding, A100
% $|\Delta|$ & 0.16 & 0.53 & 0.38 & 0.38 & 0.84 & 0.30 & 0.07 & 0.76 & 0.43 & 0.84 \\

% % Threshold-based, A100
% $|\Delta|$ & 0.08 & 0.08 & 0.60 & 0.31 & 0.00 & 0.45 & 0.61 & 1.06 & 0.40 & 1.06 \\
\begin{table}[t]
\centering
\caption{Accuracy discrepancies across GPU types under identical decoding methods and prompt templates, evaluated in float32 precision. $|\Delta|$  denotes the absolute accuracy deviation of A100 relative to L40. Avg. and Max. denote the mean and maximum $|\Delta|$  across all prompt templates, respectively.}
\label{tab:gpu_float32}
\renewcommand\arraystretch{1.2}
\resizebox{1.0\textwidth}{!}{
\setlength{\tabcolsep}{4mm}{
\begin{tabular}{lcccccccccc}
\toprule
\multirow{2}{*}{GPU} & \multicolumn{8}{c}{Template} & \multirow{2}{*}{Avg.} & \multirow{2}{*}{Max.} \\
\cmidrule(lr){2-9}
 & A & B & C & D & E & F & G & H & & \\
\hline
\rowcolor{gray!10}
L40    & 76.65 & 76.57 & 76.42 & 76.72 & 77.33 & 77.18 & 77.03 & 77.26 & -- & -- \\
\hline
A100   & 76.65 & 76.57 & 76.42 & 76.72 & 77.33 & 77.18 & 77.03 & 77.26 \\
$|\Delta|$ & 0.00 & 0.00 & 0.00 & 0.00 & 0.00 & 0.00 & 0.00 & 0.00 & 0.00 & 0.00 \\
\bottomrule
\end{tabular}
}}
\end{table}

\subsection{Experiments on multi-prompt evaluation}
\label{multi_evaluation}
In our work, we use prompt templates commonly adopted in prior studies to ensure that our evaluation reflects common practice and that the observed inconsistencies in evaluation are not due to specially designed prompts. 
To investigate the effectiveness of multi-prompt evaluation, we conduct leave-one-template-out experiments to assess the robustness of our evaluation. 
Concretely, for Templates 1–8 used in Table~\ref{tab:math_llada_results}, we exclude each template in turn and calculate Kendall’s $\tau$ between the rankings based on the average accuracy over the remaining 7 templates and all 8 templates. 
The leave-one-template-out analysis yields an average Kendall’s $\tau$ of 0.94, suggesting that averaging results across multiple templates produces a considerably more stable ranking. This further supports our recommendation to evaluate decoding methods using multiple prompt templates rather than relying on a single template.

\section{Examples of Prompt Templates and Responses}
\label{total_prompts}
\subsection{Prompt templates for mathematical and code benchmarks}
\label{label:benchmark_prompts}
Example~\ref{math_prompts} presents the prompt templates employed in our mathematical benchmark evaluation. Specifically, Templates 1–6 are designated for GSM8K, and Templates 7–8 for MATH500, and the results in Table~\ref{tab:math_llada_results} and Table~\ref{tab:math_llada1.5_results} are obtained using these templates.
All templates are sourced from lm-evaluation-harness, except for Template 6, which is newly designed in this work. Example~\ref{code_prompts} presents the prompt templates employed in code benchmark evaluation. 
Specifically, Templates 1–3 are designated for HumanEval, and Templates 4–6 for MBPP, and the results in Table~\ref{tab:code_llada_results} and Table~\ref{tab:code_llada1.5_results} are obtained using these templates. All templates are adapted from lm-evaluation-harness, with minor modifications to enhance template diversity.

\subsection{Near-identical prompt templates}
\label{label:prompt_subtle}
Example~\ref{prompt_subtle} presents a set of near-identical prompt templates constructed for the GSM8K benchmark evaluation, where each template differs from the others only in individual words or characters.
For clarity, characters that differ from the first prompt are highlighted in red.
Table~\ref{tab:prompt} reports the accuracy of different parallel decoding strategies across this set of prompt templates.

% \subsection{Effect of minor prompt variations on performance evaluation}
% To examine how minor prompt variations affect model evaluation outcomes, we construct 8 prompt templates for the GSM8K benchmark, as illustrated in Example~\ref{prompt_subtle}. 
% All templates are identical in structure and output format, where the only difference lies in a few characters of the instruction. 
% Specifically, the variations include the insertion of prefix symbols (e.g., \texttt{'}, \texttt{**}, \texttt{\$}, \texttt{??}), modifications to the imperative phrasing 
% (e.g., \textit{please}, \textit{let's}), and the substitution of a single content 
% word (\textit{question} for \textit{math problem}). 
% This design allows us to attribute any observed difference in evaluation 
% results directly to the surface-level wording of the prompt. 
% For clarity, we highlight in red the differing characters of each variant relative to the first prompt.

\subsection{Semantically equivalent prompt templates}
\label{label:prompts_step}
Example~\ref{prompts_large} presents a set of semantically equivalent prompt templates constructed for the GSM8K benchmark evaluation, where each template 
maintains identical semantics while varying in lexical and stylistic expression. Table~\ref{tab:prompt_steps} reports the accuracy and number of decoding steps of each parallel decoding strategy across this set of prompt templates.

\subsection{Chain of thought prompt templates}
\label{label:prompt_cot}
Example~\ref{prompts_cot} presents a set of prompt templates for the GSM8K benchmark, where the prompts have identical task instructions and output format requirements, differing only in the chain-of-thought elicitation phrase, ranging from no explicit reasoning instruction to increasingly detailed step-by-step guidance.

\begin{promptbox}[math_prompts]{Prompt templates for GSM8K and Math500 benchmarks}

\ttfamily
% \label{math_prompts}
\textcolor{blue}{\textbf{Template 1:}}\\
Q: \{\{question\}\}

A:

\vspace{4pt}\hrule\vspace{4pt}

\textcolor{blue}{\textbf{Template 2:}}\\
Q: \{\{question\}\}\\
A: Let's think step by step.

\vspace{4pt}\hrule\vspace{4pt}

\textcolor{blue}{\textbf{Template 3:}}\\
Question: \{\{question\}\}\\
Answer:

\vspace{4pt}\hrule\vspace{4pt}

\textcolor{blue}{\textbf{Template 4:}}\\
Question: \{\{fewshot\_question\_1\}\}\\
Answer: \{\{fewshot\_answer\_1\}\}\\

Question: \{\{fewshot\_question\_2\}\}\\
Answer: \{\{fewshot\_answer\_2\}\}\\

Question:  \{\{fewshot\_question\_3\}\}\\
Answer: \{\{fewshot\_answer\_3\}\}\\

Question: \{\{question\}\}\\
Answer:

\vspace{4pt}\hrule\vspace{4pt}

\textcolor{blue}{\textbf{Template 5:}}\\
Question: \{\{fewshot\_question\_1\}\}\\
Answer: \{\{fewshot\_answer\_1\}\}\\

Question: \{\{fewshot\_question\_2\}\}\\
Answer: \{\{fewshot\_answer\_2\}\}\\

Question: \{\{fewshot\_question\_3\}\}\\
Answer: \{\{fewshot\_answer\_3\}\}\\

Question: \{\{fewshot\_question\_4\}\}\\
Answer:\{\{fewshot\_answer\_4\}\}\\

Question: \{\{fewshot\_question\_5\}\}\\
Answer: \{\{fewshot\_answer\_5\}\}\\

Question: \{\{question\}\}\\
Answer:

\vspace{4pt}\hrule\vspace{4pt}

\textcolor{blue}{\textbf{Template 6:}}\\
Solve the following math problem step by step.

IMPORTANT OUTPUT FORMAT:\\
- Your final line must be exactly: The answer is <NUMBER>\\
- Output ONLY the number on the final line\\
\hspace*{1em}(no '\$', no commas, no units, no extra words).\\
- Put the final line on its own line.

Problem:\\
\{\{question\}\}

\vspace{4pt}\hrule\vspace{4pt}

\textcolor{blue}{\textbf{Template 7:}}\\
Problem:
\{\{question\}\}\\

Solution:
\vspace{4pt}\hrule\vspace{4pt}

\textcolor{blue}{\textbf{Template 8:}}\\
Problem:
\{\{fewshot\_question\_1\}\}\\

Solution:
\{\{fewshot\_answer\_1\}\}\\

Problem:
\{\{fewshot\_question\_2\}\}\\

Solution:
\{\{fewshot\_answer\_2\}\}\\

Problem:
\{\{fewshot\_question\_3\}\}\\

Solution:
\{\{fewshot\_answer\_3\}\}\\

Problem:
\{\{fewshot\_question\_4\}\}\\

Solution:
\{\{fewshot\_answer\_4\}\}\\

Problem:
\{\{question\}\}\\

Solution:

\end{promptbox}
\begin{promptbox}[code_prompts]{Prompt templates for HumanEval and  MBPP benchmarks}
\ttfamily
\textcolor{blue}{\textbf{Prompt 1:}}\\
\{\{Question\}\}\\
\vspace{4pt}\hrule\vspace{4pt}

\textcolor{blue}{\textbf{Prompt 2:}}\\
You are a Python coding assistant. Complete the function.

\{\{Question\}\}\\
\vspace{4pt}\hrule\vspace{4pt}

\textcolor{blue}{\textbf{Prompt 3:}}\\
You are a Python coding assistant. Follow the docstring and write the correct implementation.

\{\{Question\}\}\\

\vspace{4pt}\hrule\vspace{4pt}

% \textcolor{blue}{\textbf{Prompt D:}}\\
% You are a Python coding assistant. Produce a concise and correct function completion.

% \{\{Question\}\}\\

\textcolor{blue}{\textbf{Prompt 4:}}\\
You are an expert Python programmer, and here is your task: \{\{Question\}\} Your code should pass these tests:

\{\{test\_list[0]\}\} \\
\{\{test\_list[1]\}\} \\
\{\{test\_list[2]\}\} \\
\relax[BEGIN]

\vspace{4pt}\hrule\vspace{4pt}
\textcolor{blue}{\textbf{Prompt 5:}}\\
You are an expert Python programmer. Read the programming problem below and produce a correct Python solution: \{\{Question\}\} Your code should pass these tests:

\{\{test\_list[0]\}\} \\
\{\{test\_list[1]\}\} \\
\{\{test\_list[2]\}\} \\
\relax[BEGIN]

\vspace{4pt}\hrule\vspace{4pt}
\textcolor{blue}{\textbf{Prompt 6:}}\\
You are an expert Python programmer. Your goal is to implement a Python function that satisfies the specification and tests: \{\{Question\}\} Your code should pass these tests:

\{\{test\_list[0]\}\} \\
\{\{test\_list[1]\}\} \\
\{\{test\_list[2]\}\} \\
\relax[BEGIN]

% \vspace{4pt}\hrule\vspace{4pt}
% \textcolor{blue}{\textbf{Prompt H:}}\\
% You are an expert Python programmer. Solve the following task as Python code: \{\{Question\}\} Your code should pass these tests:

% \{\{test\_list[0]\}\} \\
% \{\{test\_list[1]\}\} \\
% \{\{test\_list[2]\}\} \\
% \relax[BEGIN]

\end{promptbox}
\begin{promptbox}[prompt_subtle]{Near-identical prompt templates}
% \label{tab:prompts_mini}
\ttfamily
\textcolor{blue}{\textbf{Prompt A:}}\\
Solve the following math problem step by step.\\
\\
IMPORTANT OUTPUT FORMAT:\\
- Your final line must be exactly: The answer is <NUMBER>\\
- Output ONLY the number on the final line (no \$, no commas, no units, no extra words).\\
- Put the final line on its own line.\\
\\
Problem: \{\{question\}\}

\vspace{4pt}\hrule\vspace{4pt}

\textcolor{blue}{\textbf{Prompt B:}}\\
\textcolor{red}{'}Solve the following math problem step by step.\\
\\
IMPORTANT OUTPUT FORMAT:\\
- Your final line must be exactly: The answer is <NUMBER>\\
- Output ONLY the number on the final line (no \$, no commas, no units, no extra words).\\
- Put the final line on its own line.\\
\\
Problem: \{\{question\}\}

\vspace{4pt}\hrule\vspace{4pt}

\textcolor{blue}{\textbf{Prompt C:}}\\
\textcolor{red}{' }Solve the following math problem step by step.\\
\\
IMPORTANT OUTPUT FORMAT:\\
- Your final line must be exactly: The answer is <NUMBER>\\
- Output ONLY the number on the final line (no \$, no commas, no units, no extra words).\\
- Put the final line on its own line.\\
\\
Problem: \{\{question\}\}

\vspace{4pt}\hrule\vspace{4pt}

\textcolor{blue}{\textbf{Prompt D:}}\\
\textcolor{red}{** }Solve the following math problem step by step\\
\\
IMPORTANT OUTPUT FORMAT:\\
- Your final line must be exactly: The answer is <NUMBER>\\
- Output ONLY the number on the final line (no \$, no commas, no units, no extra words).\\
- Put the final line on its own line.\\
\\
Problem: \{\{question\}\}

\vspace{4pt}\hrule\vspace{4pt}

\textcolor{blue}{\textbf{Prompt E:}}\\
\textcolor{red}{\$ }Solve the following math problem step by step\\
\\
IMPORTANT OUTPUT FORMAT:\\
- Your final line must be exactly: The answer is <NUMBER>\\
- Output ONLY the number on the final line (no \$, no commas, no units, no extra words).\\
- Put the final line on its own line.\\
\\
Problem: \{\{question\}\}

\vspace{4pt}\hrule\vspace{4pt}

\textcolor{blue}{\textbf{Prompt F:}}\\
\textcolor{red}{?? }Solve the following math problem step by step\\
\\
IMPORTANT OUTPUT FORMAT:\\
- Your final line must be exactly: The answer is <NUMBER>\\
- Output ONLY the number on the final line (no \$, no commas, no units, no extra words).\\
- Put the final line on its own line.\\
\\
Problem: \{\{question\}\}

\vspace{4pt}\hrule\vspace{4pt}

\textcolor{blue}{\textbf{Prompt G:}}\\
\textcolor{red}{please }solve the following math problem step by step\\
\\
IMPORTANT OUTPUT FORMAT:\\
- Your final line must be exactly: The answer is <NUMBER>\\
- Output ONLY the number on the final line (no \$, no commas, no units, no extra words).\\
- Put the final line on its own line.\\
\\
Problem: \{\{question\}\}

\vspace{4pt}\hrule\vspace{4pt}

\textcolor{blue}{\textbf{Prompt H:}}\\
\textcolor{red}{let's }solve the following math problem step by step\\
\\
IMPORTANT OUTPUT FORMAT:\\
- Your final line must be exactly: The answer is <NUMBER>\\
- Output ONLY the number on the final line (no \$, no commas, no units, no extra words).\\
- Put the final line on its own line.\\
\\
Problem: \{\{question\}\}

\end{promptbox}
\begin{promptbox}[prompts_large]{Illustrative case of prompts}
\ttfamily
\label{tab:prompts_large}
\textcolor{blue}{\textbf{Prompt A:}}\\
Solve the following math problem step by step.\\
\\
IMPORTANT OUTPUT FORMAT:\\
- Your final line must be exactly: The answer is <NUMBER>\\
- Output ONLY the number on the final line (no \$, no commas, no units, no extra words).\\
- Put the final line on its own line.\\
\\
Problem:\\
\{\{question\}\}

\vspace{4pt}\hrule\vspace{4pt}

\textcolor{blue}{\textbf{Prompt B:}}\\
Problem: \{\{question\}\}\\
\\
Solve it step by step. Last line must be exactly: The answer is <NUMBER>

\vspace{4pt}\hrule\vspace{4pt}

\textcolor{blue}{\textbf{Prompt C:}}\\
You are an expert mathematician. A student has asked you the following question:\\
\\
\{\{question\}\}\\
\\
Please solve this carefully, showing each step of your reasoning.\\
At the end, state your final answer in the format: The answer is <NUMBER>

\vspace{4pt}\hrule\vspace{4pt}

\textcolor{blue}{\textbf{Prompt D:}}\\
Think step by step to solve the following problem.\\
\\
Problem: \{\{question\}\}\\
\\
Work through the problem carefully. After your reasoning, your final answer must follow this exact format:\\
The answer is <NUMBER>

\vspace{4pt}\hrule\vspace{4pt}

\textcolor{blue}{\textbf{Prompt E:}}\\
Answer the math problem below.\\
\\
\{\{question\}\}\\
\\
End your response with: The answer is <NUMBER>

\vspace{4pt}\hrule\vspace{4pt}

\textcolor{blue}{\textbf{Prompt F:}}\\
The following is a mathematical problem that requires a step-by-step solution.\\
Analyze the problem systematically and derive the answer through logical reasoning.\\
\\
Problem Statement: \{\{question\}\}\\
\\
Required output format for the final line: The answer is <NUMBER>

\vspace{4pt}\hrule\vspace{4pt}

\textcolor{blue}{\textbf{Prompt G:}}\\
Hey, can you help me solve this math problem?\\
\\
\{\{question\}\}\\
\\
Walk me through how you'd solve it, and make sure your last line says exactly: The answer is <NUMBER>

\vspace{4pt}\hrule\vspace{4pt}

\textcolor{blue}{\textbf{Prompt H:}}\\
Solve the math problem below. Show your work.\\
\\
\{\{question\}\}\\
\\
WARNING: Your response MUST end with the following line and nothing after it:\\
The answer is <NUMBER>\\
Do NOT include units, symbols, or extra text in the final line.

\end{promptbox}
\begin{promptbox}[prompts_cot]{CoT prompt templates}
\ttfamily
\label{tab:prompts_cot}
\textcolor{blue}{\textbf{Prompt A:}}\\
Solve the following math problem.\\
\\
IMPORTANT OUTPUT FORMAT:\\
- Your final line must be exactly: The answer is <NUMBER>\\
- Output ONLY the number on the final line (no \$, no commas, no units, no extra words).\\
- Put the final line on its own line.\\
\\
Problem:\\
\{\{question\}\}

\vspace{4pt}\hrule\vspace{4pt}

\textcolor{blue}{\textbf{Prompt B:}}\\
Solve the following math problem.\\
\\
\textcolor{red}{Let's think step by step.}\\
\\
IMPORTANT OUTPUT FORMAT:\\
- Your final line must be exactly: The answer is <NUMBER>\\
- Output ONLY the number on the final line (no \$, no commas, no units, no extra words).\\
- Put the final line on its own line.\\
\\
Problem:\\
\{\{question\}\}

\vspace{4pt}\hrule\vspace{4pt}

\textcolor{blue}{\textbf{Prompt C:}}\\
Solve the following math problem.\\
\\
\textcolor{red}{Reason step by step before giving the final answer.}\\
\\
IMPORTANT OUTPUT FORMAT:\\
- Your final line must be exactly: The answer is <NUMBER>\\
- Output ONLY the number on the final line (no \$, no commas, no units, no extra words).\\
- Put the final line on its own line.\\
\\
Problem:\\
\{\{question\}\}

\vspace{4pt}\hrule\vspace{4pt}

\textcolor{blue}{\textbf{Prompt D:}}\\
Solve the following math problem.\\
\\
\textcolor{red}{Work through the solution carefully step by step before giving the final answer.}\\
\\
IMPORTANT OUTPUT FORMAT:\\
- Your final line must be exactly: The answer is <NUMBER>\\
- Output ONLY the number on the final line (no \$, no commas, no units, no extra words).\\
- Put the final line on its own line.\\
\\
Problem:\\
\{\{question\}\}

\vspace{4pt}\hrule\vspace{4pt}

\textcolor{blue}{\textbf{Prompt E:}}\\
Solve the following math problem.\\
\\
\textcolor{red}{Show your reasoning and work through the solution very carefully step by step before giving the final answer.}\\
\\
IMPORTANT OUTPUT FORMAT:\\
- Your final line must be exactly: The answer is <NUMBER>\\
- Output ONLY the number on the final line (no \$, no commas, no units, no extra words).\\
- Put the final line on its own line.\\
\\
Problem:\\
\{\{question\}\}

\end{promptbox}
% \begin{tcolorbox}[
%     colback=blue!5,
%     colframe=gray!100,
%     boxrule=0.5pt,
%     arc=2pt,
%     left=5pt,
%     right=5pt,
%     top=5pt,
%     bottom=5pt,
%     title={An illustration of prompt-induced trajectory divergence},
%     label={examples_discrepancies},
%     segmentation style={solid, gray!50, line width=0.5pt}
% ]
\begin{examplebox}[breakable, label={box:discrepancies}]{An illustration of prompt-induced trajectory divergence}
\small
\textbf{Prompt A:}

Solve the following math problem step by step.

IMPORTANT OUTPUT FORMAT:
\begin{itemize}
    \item Your final line must be exactly: \texttt{The answer is <NUMBER>}
    \item Output ONLY the number on the final line (no '\$', no commas, no units, no extra words).
    \item Put the final line on its own line.
\end{itemize}

Problem:

Carla is downloading a 200 GB file. Normally she can download 2 GB/minute, but 40\% of the way through the download, Windows forces a restart to install updates, which takes 20 minutes. Then Carla has to restart the download from the beginning. How long does it take to download the file?

\vspace{2mm}

Response:

1. Calculate the time taken to download the first 40\% of the file: 
\textcolor{red}{80 GB / 2 GB/minute = 40 minutes.}

2. Calculate the time taken to download the remaining 60\% of the file: 
\textcolor{red}{120 GB} / 2 GB/minute = 60 minutes.

3. Add the time for the restart\textcolor{red}{: 20 minutes.}

\textcolor{red}{4. Sum the times for all the steps: 40 minutes + 60 minutes + 20 minutes = 120 minutes.}

The answer is \textcolor{red}{120}

\tcbline

\textbf{Prompt B:}

\textcolor{red}{**} Solve the following math problem step by step

IMPORTANT OUTPUT FORMAT:
\begin{itemize}
    \item Your final line must be exactly: \texttt{The answer is <NUMBER>}
    \item Output ONLY the number on the final line (no '\$', no commas, no units, no extra words).
    \item Put the final line on its own line.
\end{itemize}

Problem:

Carla is downloading a 200 GB file. Normally she can download 2 GB/minute, but 40\% of the way through the download, Windows forces a restart to install updates, which takes 20 minutes. Then Carla has to restart the download from the beginning. How long does it take to download the file?

Response:

1. Calculate the time taken to download the first 40\% of the file: 
\textcolor{red}{40\% of 200 GB / 2 GB/minute = 80 minutes.}

2. Calculate the time taken to download the remaining 60\% of the file: 
\textcolor{red}{60\% of 200 GB} / 2 GB/minute = 60 minutes.

3. Add the time for the restart \textcolor{red}{ and the initial download: 80 minutes + 60 minutes + 20 minutes = 160 minutes.}

The answer is \textcolor{red}{160}\textcolor{red}{.}

% \end{tcolorbox}
\end{examplebox}

\end{document}